\documentclass{article}

\usepackage[numbers]{natbib}

\usepackage[final]{neurips_2022}

\usepackage[utf8]{inputenc} 
\usepackage[T1]{fontenc}    
\usepackage[hidelinks]{hyperref}       
\usepackage{url}            
\usepackage{booktabs}       
\usepackage{amsfonts}       
\usepackage{nicefrac}       
\usepackage{microtype}      
\usepackage[dvipsnames]{xcolor}        

\usepackage{wrapfig}
\usepackage{graphicx}
\usepackage{subcaption}
\usepackage{dcolumn}
\usepackage{setspace}
\usepackage{algorithm,algpseudocode}
\usepackage[algo2e]{algorithm2e} 
\usepackage{multirow, makecell}
\usepackage{amsmath}
\usepackage{bm}
\usepackage{cleveref}
  {
      \newtheorem{assumption}{Assumption}
  }
    {
      \newtheorem{proposition}{Proposition}
  }
  
      {
      \newtheorem{corollary}{Corollary}
  }
        {
      
  }
  {
    \newtheorem{theorem}{Theorem}
  }
    {
    
  }
    {
    \newtheorem{definition}{Definition}
  }
    {
    
  }
     {
    \newtheorem{example}{Example}
  }
  
\DeclareMathOperator*{\argmax}{argmax}
\usepackage{amssymb}
\usepackage{pifont}
\newcommand{\g}[2]{#1\textsubscript{\textcolor{gray}{$\pm$#2}}}
\newcommand{\highlight}[1]{\colorbox{blue!10}{#1}}
\usepackage{color,soul}
\makeatletter
\newcommand{\printfnsymbol}[1]{%
  \textsuperscript{\@fnsymbol{#1}}%
}
\hypersetup{
    colorlinks,
    linkcolor={red},
    citecolor={blue},
    urlcolor={blue}
}
\makeatother
\title{Continual Learning In Environments With Polynomial Mixing Times}

\author{%
 Matthew Riemer\thanks{Equal Contribution \newline\textsuperscript{\rm 1}IBM Research; \textsuperscript{\rm 2}Mila, Université de Montréal; \textsuperscript{\rm 3}Massachusetts Institute of Technology} \textsuperscript{ \rm 1, 2}, 
    Sharath Chandra Raparthy\printfnsymbol{1} \!\!\textsuperscript{\rm 2},
    Ignacio Cases\textsuperscript{\rm 3},
    Gopeshh Subbaraj\textsuperscript{\rm 2},\and
    \textbf{Maximilian Puelma Touzel}\textsuperscript{\rm 2},
    \textbf{Irina Rish}\textsuperscript{\rm 2}
}

\begin{document}
\maketitle
\begin{abstract}
The mixing time of the Markov chain induced by a policy limits performance in real-world continual learning scenarios. Yet, the effect of mixing times on learning in continual reinforcement learning (RL) remains underexplored. In this paper, we characterize problems that are of long-term interest to the development of continual RL, which we call scalable MDPs, through the lens of mixing times. In particular, we theoretically establish that scalable MDPs have mixing times that scale polynomially with the size of the problem. We go on to demonstrate that polynomial mixing times present significant difficulties for existing approaches, which suffer from myopic bias and stale bootstrapped estimates. To validate our theory, we study the empirical scaling behavior of mixing times with respect to the number of tasks and task duration for high performing policies deployed across multiple Atari games. Our analysis demonstrates both that polynomial mixing times do emerge in practice and how their existence may lead to unstable learning behavior like catastrophic forgetting in continual learning settings.   
\end{abstract}
\section{Introduction}
Continual reinforcement learning (RL) \citep{crlsurvey} is an aspirational field of research confronting the difficulties of long-term, real-world applications by studying problems of increasing scale, diversity, and non-stationarity. The practical requirement for researchers to work on problems of reasonable complexity in the short-term presents a meta-challenge: choosing the right small-scale problems so that the approaches we develop scale up to the use cases of the future. Here, we address this meta-challenge by formalizing RL problems that vary in size and by analyzing the scaling behavior of popular RL algorithms.  In particular, we analyze on the often ignored \textit{mixing time} that expresses the amount of time until the agent-environment dynamics converge to some stationary behaviour.

Towards this end, we specifically make the following contributions in this work: 
\begin{enumerate}
    \item \textbf{Scalable MDPs:} We propose the formalism of \textit{scalable MDPs} in Definition \ref{def:scalableMDP} to characterize an abstract class of MDPs where the MDPs within the class are differentiated based on a changing scale parameter. Understanding the influence of this scale parameter on learning can promote better understanding of the meta-challenge of extrapolating our results on small scale problems to large scale problems of the same class. 
    \item \textbf{%
    Polynomial Mixing Times:} Theorem \ref{theorem:scalablemixing} establishes the key result of this paper that as any scalable MDP is scaled, its mixing time must grow polynomially as a function of the growing state space. This has major implications for regret analysis in these MDPs.
    \item \textbf{Myopic Bias During Scaling:} We demonstrate in Corollaries \ref{corollary:mc}, \ref{corollary:boot}, and \ref{corollary:improvement} that Theorem \ref{theorem:scalablemixing} implies traditional approaches to RL cannot efficiently scale to problems of large size without experiencing myopic bias in optimization that slows down learning.  
    \item \textbf{Empirical Analysis of Mixing for Continual RL:} We back up our theory with empirical analysis of mixing time scaling in continual RL settings based on the Atari and Mujoco benchmarks. Our analysis provides insight into why agents experience significant catastrophic forgetting or other optimization instability when learning in these domains. We hope that our analysis into the fundamental driver of these issues can open the door for more successful and principled approaches to continual RL domains like these moving forward. 
\end{enumerate}
We refer readers to Appendix \ref{app:proofs} for detailed proofs of all propositions and theorems in our paper.
\vspace{-1mm}
\section{Formalizing Our Continual RL Setting}
\vspace{-1mm}
Aiming to characterize a broad range of settings, we focus on the formulation of RL in continuing environments. As explained in \citet{crlsurvey}, not only are supervised learning and episodic RL special cases of RL in continuing environments, but so are their non-stationary variants (continual supervised learning and continual episodic RL), despite violating the stationarity assumptions of their root settings. Indeed, approaches that do not acknowledge the more general RL in continuing environments setting exhibit myopic bias in their optimization when faced with non-stationarity \citep{crlsurvey}. 

Unfortunately, the popular discounted reward setting inserts the very same kind of myopic bias in optimization that we would like to avoid \citep{rlearning}. 
As typically implemented, discounting does not correspond to the maximization of any objective function over a set of policies \citep{stopdiscounting} and the policy gradient is not the gradient of any function \citep{notagradients}. These fundamental issues do not resolve as the discount factor approaches 1 \citep{stopdiscounting} and discounting does not influence the ordering of policies, suggesting it likely has no role to play in the definition of the control problem \citep{SuttonNewBook}. %
In contrast, the average reward per step objective, which explicitly includes an average over the stationary distribution, avoids inducing myopic biases and hence is well-suited for continual RL problems \citep{SuttonNewBook,crlsurvey}. %
\vspace{-1mm}
\subsection{Average Reward RL in Continuing Environments}
\vspace{-1mm}
RL in continuing environments is typically formulated using a finite, discrete-time, infinite horizon Markov Decision Process (MDP) \citep{puterman1994markov, SuttonNewBook}, which is a tuple $ {\cal M} = \langle {\cal S}, {\cal A}, T, R \rangle $, where ${\cal S}$ is the set of states, ${\cal A}$ is the  set of actions, $R: {\cal S} \times {\cal A}\rightarrow [0,R^\textrm{max}]$ is the reward function, and $T:{\cal S} \times {\cal S} \times {\cal A} \rightarrow \left[0,1\right]$ is the environment transition probability function. At each time step, the learning agent perceives a state $s \in {\cal S}$ and takes an action $a \in {\cal A}$ drawn from a policy $\pi : {\cal S} \times {\cal A} \rightarrow [0,1]$ with internal parameters $\theta \in \Theta$. The agent then receives a reward $R(s,a)$ and with probability $T(s'|s,a)$ enters next state $s'$. Markov chains may be periodic and have multiple recurrent classes, but optimality is difficult to define in such cases \citep{learningandplanning}, making the following assumption necessary for analysis:
\vspace{-1mm}
\begin{assumption} \label{ergodicassumption}
All stationary policies are aperiodic and unichain, meaning they give rise to a Markov chain with a single recurrent class %
that is recurrent in the Markov chain of every policy.\footnote{This corresponds to what is called an ergodicity assumption for all stationary policies in \citep{SuttonNewBook}.}
\end{assumption}
\vspace{-1mm}
Any RL problem may be modified such that Assumption \ref{ergodicassumption} holds by adding an arbitrarily small positive constant $\epsilon$ to all transition probabilities in $T(s'|s,a)$ and renormalizing in which case the effect on the objective of each stationary policy is $O(\epsilon)$ \citep{bertsekas98}. 
An important corollary to Assumption \ref{ergodicassumption} is that the \textit{steady-state distribution} $\mu^\pi$ induced by the policy $\pi$ is independent of the initial state:

\vspace{-1mm}
\begin{corollary} \label{cor1}
All stationary policies $\pi$ induce a unique steady-state distribution $\mu^\pi(s) = \lim_{t \rightarrow \infty} P^\pi( \textcolor{black}{s_t = s}|s_0 )$ that is independent of the initial state such that $\sum_{s \in \cal S} \mu^\pi(s) \sum_{a \in \cal A} \pi(a|s) T(s'|s,a) = \mu^\pi(s') \;\; \forall s' \in \cal S$.
\end{corollary}
\vspace{-1mm}
Corollary 1 implies that the long-term rewards of any $\pi$ will be independent of the current state. As such, the average reward per step objective $\rho(\pi)$ can be defined independently of its starting state \citep{SuttonNewBook}:
\begin{align} \label{averagereturn}
    \rho(\pi) &:= \lim_{h \rightarrow \infty} \frac{1}{h} \sum_{t=1}^h \mathbb{E}_{\pi} \bigg[ R(s_t,a_t) \bigg] = \lim_{t \rightarrow \infty} \mathbb{E}_{\pi} \bigg[ R(s_t,a_t) \bigg] \nonumber\\ 
    &= \sum_{s \in \cal S}  \mu^\pi(s) \sum_{a \in \cal A} \pi(a|s) R(s,a)\;.
\end{align}
Computing the average reward with the last expression is limited by the amount of time the Markov chain induced by the policy $T^\pi(s'|s) = \sum_{a \in \cal A} \pi(a|s) T(s'|s,a)$ needs to be run for before reaching the steady-state distribution $\mu^\pi(s)$. This amount of time is referred to in the literature as the mixing time of the induced Markov chain. We denote $t_\textrm{mix}^\pi(\epsilon)$ as the \textit{$\epsilon$-mixing time} of the chain induced by $\pi$: 
\begin{equation} \label{eq:mixingtimes}
\begin{split}
    t_\textrm{mix}^\pi(\epsilon) &:= \min \Big\{ h \;\Big| \max_{s_0 \in \cal S} d_\textrm{TV}\big(\textcolor{black}{P^\pi(s_h=\cdot|s_0), \mu^\pi(\cdot)}\big) \leq \epsilon \Big\} \nonumber
\end{split}
\end{equation}
where $d_\textrm{TV}$ is the total variation distance between the two distributions. The so-called \textit{conventional mixing time} is defined as $t_\textrm{mix}^\pi \equiv t_\textrm{mix}^\pi(1/4)$. The conventional mixing time only gives insight about distributional mismatch with respect to the steady-state distribution, which led \citep{kearns2002near} to introduce the notion of a mismatch with respect to the reward rate. The $\epsilon$\textit{-return mixing time} is a measure of the time it takes to formulate an accurate estimate of the true reward rate. 
More formally, if we denote the $h$-step average undiscounted return starting from state $s$ as $\rho(\pi, s, h)$, then we define the $\epsilon$\textit{-return mixing time} as:
\begin{equation} \label{eq:mixingtimes-ret}
t_{\textrm{ret}}^\pi(\epsilon) := \min \Big\{ h \;\Big| |\rho(\pi, s_0, h')- \rho(\pi) | \leq \epsilon,  \ \ \forall s_0 \in \mathcal{S} \ \textrm{and} \ \forall h' \geq h \Big\}
\end{equation}
We will come back to this definition when we present our experiments. As emphasized in \citet{mixingtimesbook}, however, one should not get bogged down in the use-case specific definitions of mixing time (we  prove their equivalent scaling behaviour in fact, \textit{c.f.} Proposition \ref{prop:polynomialmixing} ). %
Rather, it is their shared property of being determined by both a policy $\pi$ and environment $\cal M$, and not simply a property of the environment itself, that is important. Thus the environment's contribution can be assessed only through mixing times obtained from extreme policies such as the optimal policy, $t_\textrm{mix}^{\pi^*}$.

An alternative approach for quantifying mixing is by considering the structure of the transition matrix $T^\pi(s'|s)$ induced by $\pi$. It is well known that the mixing properties of a Markov chain are governed by the spectral gap derived from the eigenvalues of the matrix $T^\pi(s'|s)$. 
Unfortunately, it is difficult to reason about the spectral gap of a class of MDPs directly. 
Towards this end, one useful interpretation of a Markov chain is as a random walk over a graph $\mathcal{G}(\pi,T)$ with vertex set $\cal S$ and edge set $\{(s,s')\}$, for all $s$ and $s'$ satisfying $ T^\pi(s'|s) + T^\pi(s|s') > 0$. The \textit{diameter} $D^\pi$ of the Markov chain induced by $\pi$ is thus the diameter or maximal graph distance of $\mathcal{G}(\pi,T)$. It is defined using the \textit{hitting time} $t_\textrm{hit}^\pi(s_1|s_0)$, the first time step in which $s_1$ is reached following the Markov chain $T^\pi(s'|s)$ from $s_0$, 
\begin{equation} \label{eq:diameters}
    D^\pi := \max_{s_0, s_1 \in \mathcal{S}} \mathbb{E}_\pi \big[ t_\textrm{hit}^\pi(s_1|s_0) \big] \;\;;\;\; D^* := \min_\pi  D^\pi \;.
\end{equation}
$D^*$ then denotes the \textit{minimum diameter} achieved by any policy. 
All MDPs that follow Assumption \ref{ergodicassumption} have finite diameter $D^\pi$ for all policies \citep{jaksch2010near}. 
\citet{mixingtimesbook} (eq. 7.4) established the relationship, $2t_\textrm{mix}^{\pi} \geq D^\pi$, so that the diameter provides a lower bound for the mixing time and thus serves as a very relevant quantity when conducting mixing time analyses.
\vspace{-1mm}
\subsection{The Role Of Tasks And Non-stationarity } \label{sec:taskdefinition}

\vspace{-1mm}
The preceding single-task formulation of RL in continuing environments can, in fact, be used to formulate continual RL, which typically makes notions of multiple tasks and non-stationarity explicit. 
This is indeed a useful construct because
arbitrary non-stationarity precludes any consistent signal to learn from, and thus further assumptions about the environment structure must be made to make progress \citep{crlsurvey}. In this paper, we consider tasks as sub-regions of the total MDP (which alternatively can be thought of as independent MDPs connected together) and we assume the transition dynamics between and within tasks are both stationary. As highlighted in \cite{crlsurvey}, this setting is quite general as it is capable of modeling many continual RL problems. For example, it easily extends to partially observable variants: if the task component is not directly observed, the problem appears non-stationary from the perspective of an agent that learns from observations of only the within-task state. 
\textcolor{black}{
That said, the continual learning literature demonstrates that optimization issues are generally experienced whether task labels are observed or not \citep{delange}. Our analysis of mixing times in this paper is an attempt to understand these difficulties even when the total MDP is fully observable and stationary. 
}

\begin{figure}
    \centering
    \includegraphics[width=0.9\textwidth]{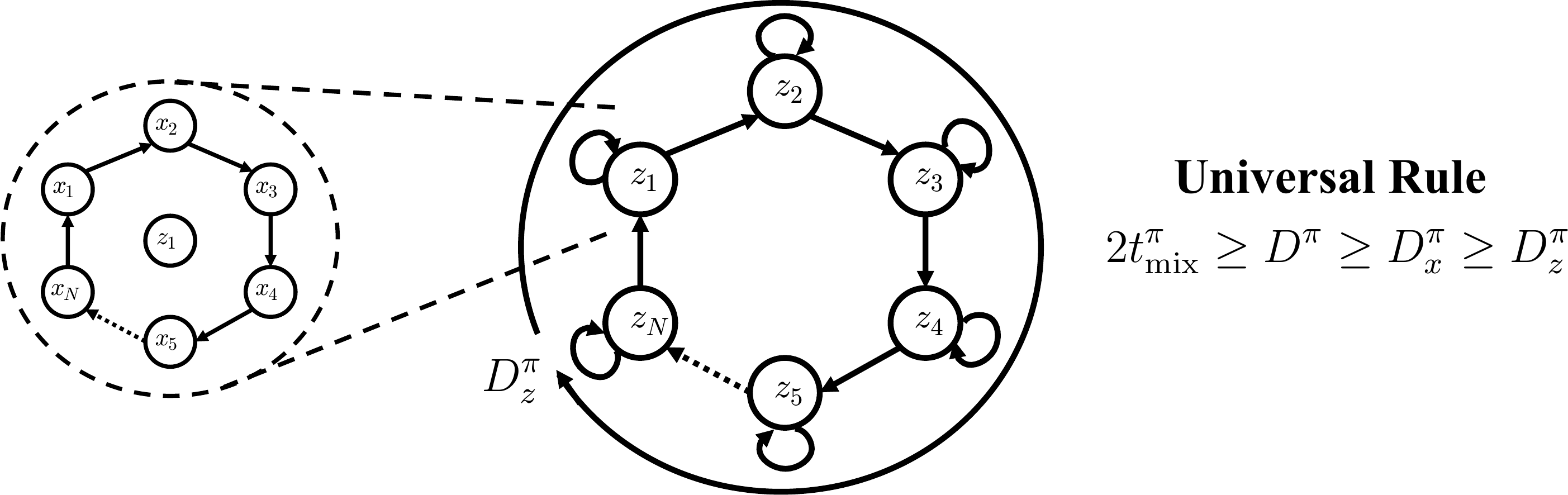}
    \caption{\textbf{Continual RL Setting:} The state space $s \in \mathcal{S}$ is decomposed as $s=[x,z]$ where $z \in \mathcal{Z}$ is the task and $x \in \mathcal{X}_z$ is the within-task state. The mixing time, $t^\pi_\textrm{mix}$, is lower-bounded by the diameter over the full state space $D^\pi$. The latter is lower-bounded by the diameter over the within-task state space $D^\pi_x$, which in turn is lower-bounded by the diameter over the space of tasks, $D^\pi_z$.}
    \label{fig:cont_rl}
    \vspace{-4mm}
\end{figure}

To help the reader picture $t^\pi_\textrm{mix}$ in continual RL, we offer the example depicted in Figure \ref{fig:cont_rl}. Here, the agent's state space $\mathcal{S}$ is decomposed into a task component, $z \in \mathcal{Z}$ and a within-task component, $x \in \mathcal{X}_z$. For a policy, $\pi$, the pair of Markov chains induced by marginalizing over $x$ and over $z$ have diameters $D_z^\pi$ and $D_x^\pi$, respectively. 
Using the result connecting mixing times and diameters here, we can establish a universal rule for problems of this type: $2t_\textrm{mix}^{\pi} \geq D^\pi \geq D_x^\pi \geq D_z^\pi$, where $D^\pi$ is the diameter over the entire state space. This rule highlights the intimate connection between mixing times and continual RL: task locality and bottleneck structure inherently lead to environments with correspondingly high mixing times due to high minimum diameters between states in different tasks.  

\vspace{-1mm}
\section{Scalable MDPs} 
\begin{figure}[tbph!]
    \centering
   \includegraphics[width=0.85\linewidth]{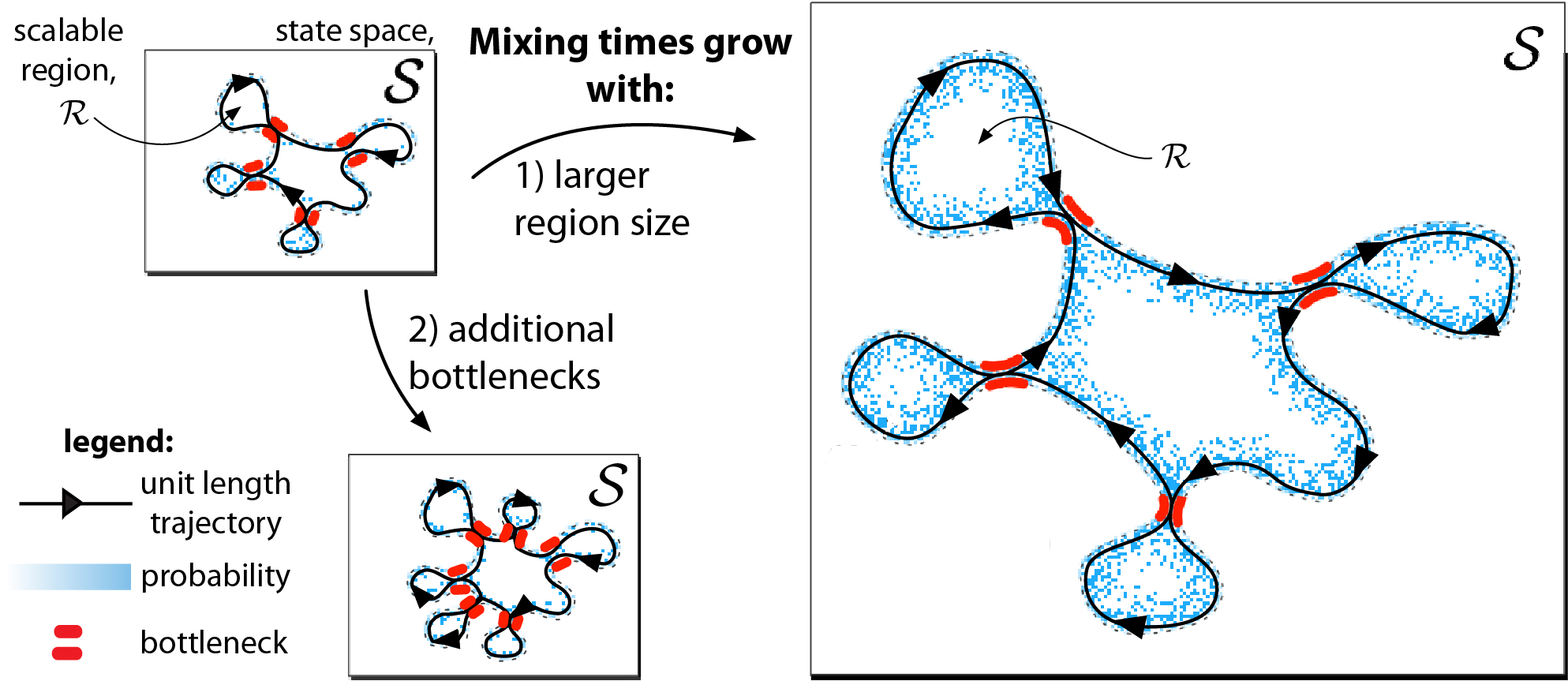}
   
    \caption{\textcolor{black}{\small{
    \textbf{Mixing times grow as MDPs are scaled up.} \textit{Top left}: A more general version of the continual RL setting shown in Figure \ref{fig:cont_rl}, where individual tasks correspond to regions, $\mathcal{R}$, of the state space, $\mathcal{S}$, connected through bottlenecks (see legend). 
    An example of the possible steady-state probability of $\pi^*$ is shown in black gradient. 
    The mixing time of an MDP can grow by increasing its diameter (number of equidistant arrow heads) via (1) scaling that increases the size of visited regions $\mathcal{R}$ and thus the expected residence time $t^{\pi^*}_\mathcal{R}$ of $\pi^*$ (right) and/or (2) scaling that increases the number of bottlenecks between regions of the state space (bottom left). 
    }
    }
        }
    \label{fig:scaled}
\end{figure}

\vspace{-1mm}
\textcolor{black}{
In this section, we formalize the notion of scaling in the context of MDPs and provide intuition about the effect of this scaling with the aid of the schematic shown in \Cref{fig:scaled}. 
We assume MDPs can be described in terms of an $n$-dimensional parameter vector $\bm{q}\in\mathbb{R}^n$, and that some subset of these dimensions are the parameters of interest that will be scaled up. We would like to analyze the effect of this scaling on mixing times.\footnote{\textcolor{black}{Scalar parameters with discrete domains are incorporated by embedding them into $\mathbb{R}$.}}  
With reference to \Cref{fig:cont_rl}, these parameters can be non-spatial such as the number of tasks, or spatial such as the size of an individual task.
More generally and with reference to \Cref{fig:scaled}, the effect of scaling up parameters can contribute to longer mixing times by adding bottleneck states (akin to increasing the number of tasks) and increasing the size of the regions generated by a bottleneck (akin to increasing the size of each task). 
How parameters are scaled is specified by a scaling function, $\sigma$ controlled by scaling parameter, $\nu$.\footnote{\textcolor{black}{Formally, this is a scaling deformation, $\sigma:\mathbb{R}^n\times\mathbb{R}\to\mathbb{R}^n$ parameterized by an order parameter $\nu \in \mathbb{R}$ that takes any $\bm{q}_0$ to $\bm{q}_\nu=\sigma(\bm{q}_0,\nu)$, with $\sigma(\cdot,0)$ as the identity map. Proportional scaling has $q_{\nu,i}\propto\nu$ for all $i$ indexing scaled parameters such that $\bm{q}_{\nu}=\bm{q}_0+\nu\Delta \bm{q}$, with $\Delta\bm{q}\in\mathbb{R}^n$ giving the rates of linear growth in $\bm{q}$ with $\nu$ up from $\bm{q}_0$.}} 
All MDPs accessible through $\sigma$ by varying $\nu$ form a family of MDPs, $\mathbb{C}_\sigma=\{\mathcal{M}_\nu\}$, where $\mathcal{M}_\nu$ is the MDP specified by $\bm{q}_\nu$. 
In particular, we consider \textit{proportional scaling} functions, for which a subset of elements of $\bm{q}$ are scaled linearly with $\nu$. 
In our experiments, we show results for proportional scaling of two MDP parameters central to continual RL: 
the number of tasks and the time between task switches.}

\textcolor{black}{As the MDP scales up according to $\sigma$, the overall state space size will grow with $|\mathcal{S}|\to\infty$ as $\nu\to\infty$. Regions within the state space can also grow and the steady-state probability on them from $\pi$ may change as a result.}
A region $\mathcal{R}\subseteq\mathcal{S}$ is a connected subset of states with steady-state probability $\mu^\pi(\mathcal{R})=\sum_{s \in \mathcal{R}} \mu^\pi(s)$. 
The boundary of $\mathcal{R}$ is a subset $\partial\mathcal{R} \subseteq \mathcal{R}$ with states having finite probability of transitioning to at least one state that is outside of $\mathcal{R}$, $\sum_{s' \in \mathcal{S}\setminus\mathcal{R}} T^\pi(s'|s) > 0 \;\; \forall s \in \partial \mathcal{R}$.
\textcolor{black}{
We denote the scalable regions and boundaries $\mathcal{R}_\nu$ and $\partial\mathcal{R}_\nu$, respectively.
If the size of a region's boundary grows faster than the region's interior, there is a finite state space size  
at which the region no longer has an interior.} %
In that case, the problem type is of bounded complexity and not relevant to the development of scaling friendly algorithms.
We are thus interested in problems where scalable regions can maintain an interior as they are scaled (as measured by steady-state probability) in the limit of large $\nu$. \textcolor{black}{See Appendix \ref{app:proofs} for a precise formulation leading to the following definition:}

\begin{definition} \label{def:scalableMDP}
A \textbf{scalable MDP} is a family of MDPs \textcolor{black}{$\mathbb{C}_\sigma=\{\mathcal{M}_\nu\}$ 
arising from a proportional scaling function $\sigma$ 
satisfying the property that 
there exists an initial scalable region $\mathcal{R}_0$ with finite interior, $\mu^{\pi^*}(\partial\mathcal{R}_0) < \mu^{\pi^*}(\mathcal{R}_0)$, that scales so that $\mu^{\pi^*}(\partial\mathcal{R}_\nu) < \mu^{\pi^*}(\mathcal{R}_\nu)$ %
as $\nu\to\infty$ and thus $|\mathcal{S}|\to\infty$.}%
\end{definition}
\vspace{-1mm}
\section{Polynomial Mixing Times}
\vspace{-1mm}
Prior work analyzing the scaling of mixing times makes the practical distinction between \textit{slow mixing} Markov chains that scale exponentially with a size parameter and \textit{rapid mixing} Markov chains that scale at most polynomially. Since the learning speed of any algorithm is lower bounded by the mixing time of the optimal policy %
\citep{kearns2002near}, environments with a slow mixing Markov chain induced by the optimal policy may very well be outside the practical reach of RL algorithms. In this section, we would like to further characterize mixing times over the space of practically solvable problems of sufficient complexity to still be of interest for real-world applications.

For continuing environments, the tightest known lower bound satisfying Assumption \ref{ergodicassumption}\footnote{The bias span is only lower for weakly communicating MDPs violating Assumption \ref{ergodicassumption} so that $D^*=\infty$.} has $H$-step regret $\textrm{Regret}(H) \in \Omega(\sqrt{D^{\pi^*}|\mathcal{S}||\mathcal{A}|H}) \subseteq \Omega(\sqrt{D^*|\mathcal{S}||\mathcal{A}|H})$ \citep{jaksch2010near}. %
Additionally, because we know that $D^* \geq \log_{|\mathcal{A}|}(|\mathcal{S}|)-3$ \citep{jaksch2010near}, we can bound our regret in the general case as $\textrm{Regret}(H) \in \tilde{\Omega}(\sqrt{|\mathcal{S}||\mathcal{A}|H})$. %
Few practical problems are simple enough to exhibit diameters with such logarithmic contributions. More typical and thus of greater interest are diameters (and mixing times) having polynomial scaling in $|\mathcal{S}|$. 
Focusing on polynomial scaling is thus warranted, and provides more stringent lower bounds on regret. We thus consider the following definition: %

\begin{definition} \label{def:polynomialmixing}
A set or family of MDPs $\mathbb{C}$ %
has a \textbf{polynomial mixing time} if the environment mixing dynamics contributes a $\Omega(|\mathcal{S}|^k)$ multiplicative increase for some $k>0$ to the intrinsic lower bound on regret %
as $|\mathcal{S}| \rightarrow \infty \;\; \forall \mathcal{M} \in \mathbb{C}$.
\end{definition}
An immediate utility of Definition \ref{def:polynomialmixing} is that it subsumes the diversity of different diameter and mixing time definitions by explicitly stating their equivalence
in scaling with respect to the state space size: 

\begin{proposition} \label{prop:polynomialmixing}
If all MDPs $\mathcal{M}$ within the subclass of MDPs $\mathbb{C}$ have $t_\textnormal{ret}^{\pi^*}$, $t_\textnormal{ces}^{\pi^*}$, $t_\textnormal{mix}^{\pi^*}$, $D^{\pi^*}$, or $D^* \in \Omega(|\mathcal{S}|^k)$ for some $k>0$ we can say that $\mathbb{C}$ has a polynomial mixing time.\footnote{The $\epsilon$-return mixing time $t_\textrm{ret}^\pi(\epsilon)$ is for assessing mixing from the perspective of accumulated rewards, whereas the $\epsilon$-Cesaro mixing time $t_\textrm{ces}^\pi(\epsilon)$ is for periodic problems that do not converge in the limit from Corollary \ref{cor1}. See Appendix \ref{app:proofs} for additional details.}
\end{proposition}
We can thus hereon focus on formulating scaling in MDPs through their state space size.
One way to understand how a \textcolor{black}{scalable region, $\mathcal{R}_\nu$}, contributes to the mixing time is through its \textit{residence time}, $t^{\pi^*}_\mathcal{R}$, \textit{i.e.} the average time that $\pi^*$ spends in $\mathcal{R}$ during a single visit.
Using results from the theory of bottleneck ratios in Markov chains \citep{mixingtimesbook}, we show that if scaling the MDP leads to  $t^{\pi^*}_\mathcal{R}$ increasing by a polynomial factor in $|\mathcal{S}|$, then the mixing time also increases at a polynomial rate:

\begin{proposition} \label{prop:timespent}
\textcolor{black}{Any scalable MDP $\mathbb{C}_\sigma$ exhibits a polynomial mixing time if there exists a scalable region $\mathcal{R}_\nu$ 
such that $\mathbb{E}_{\mu^{\pi^*}}[t^{\pi^*}_{\mathcal{R}_\nu}] \in \Omega(|\mathcal{S}|^{k})$ for some  $k > 0$.}
\end{proposition}

The purpose of Proposition \ref{prop:timespent} and Figure \ref{fig:scaled} are to provide intuition to readers about why scalable MDPs inherently must have polynomial mixing times. \textcolor{black}{Our paper's main result builds off Proposition \ref{prop:timespent}, to make a general statement about the set of all possible scalable MDPs:} 
\begin{theorem} \label{theorem:scalablemixing}
\textcolor{black}{(Mixing Time Scaling): \;
Any scalable MDP $\mathbb{C}_\sigma$ has a polynomial mixing time.}
\end{theorem}
\vspace{-1mm}
\section{Myopic Bias During Scaling}
\vspace{-1mm}
\label{sec:myopicbias}
Monte Carlo sampling and bootstrapping are the two primary policy evaluation frameworks for RL \citep{SuttonNewBook}. Both implicitly assume that a finite and fixed maximum frequency $f^*(\pi) \in [0,1]$ of policy improvement steps is possible for unbiased updates regardless of the problem size. We now propose three corollaries of Theorem \ref{theorem:scalablemixing} that together argue how polynomial mixing times invalidate this assumption simply because $f^*(\pi)\geq 1/t^\pi_\textrm{mix}$, and $t^\pi_\textrm{mix}$ grows over scalable MDPs. \textcolor{black}{In practice, algorithm designers cannot afford to wait until reaching the mixing time before updating their model when mixing times are high. This results in a myopic bias in the policy improvement steps taken.}

\textbf{Monte Carlo sampling:} 
In continuing environments with polynomial mixing times, this sampling procedure poses a problem when optimizing for $\rho(\pi)$. To get an unbiased estimate of $\rho(\pi)$, we must be able to sample from the steady-state distribution $\mu^\pi(s)$, which is only available after $t^\pi_\textrm{mix}$ steps in the environment.  Moreover, as demonstrated by \citet{zahavy2020unknown} in the general case where no upper bound on the mixing time is known apriori, $O(|\mathcal{S}| t^\pi_\textrm{mix})$ samples are needed to retrieve a single unbiased sample from $\mu^\pi(s)$. As such, it is clear that the length of the policy evaluation phase strongly depends on the mixing time of the current policy $\pi$ and thus the maximum frequency $f^*(\pi)$ of unbiased policy improvement steps decreases as the mixing time increases:
\vspace{-1mm}
\begin{corollary}  \label{corollary:mc}
A Monte Carlo sampling algorithm for policy $\pi$ in scalable MDP $\mathbb{C}_\sigma$ has a maximum frequency of unbiased policy improvement steps $f^*(\pi) \rightarrow 0$ as $|\mathcal{S}| \rightarrow \infty$.
\end{corollary}
\vspace{-1mm}
For scalable MDPs of significant size, Corollary \ref{corollary:mc} implies that model-free Monte Carlo methods perform unbiased updates arbitrarily slowly and that model-based Monte Carlo methods will need arbitrary amounts of compute for unbiased updates even when a true environment model is known. As such, to address scaling concerns over large horizons, bootstrapping methods based on the Bellman equation are generally recommended \citep{SuttonNewBook}. 

\textbf{Bootstrapping for Evaluation:} 
This process is referred to as \textit{iterative policy evaluation} \citep{SuttonNewBook} and is known to converge to the true $V^\pi$ in the limit of infinite evaluations of a fixed policy. However, in practice, RL algorithms constantly change their policy as they learn and only can afford partial backups during policy evaluation when applied to large-scale domains. Indeed, a foundational theoretical principle in RL is that as long as the agent constantly explores all states and actions with some probability, bootstrapping with only partial backups will still allow an agent to learn $\pi^*$ in the limit of many samples \citep{SuttonNewBook}. However, sample efficiency can still be quite poor since partial backups insert bias into each individual policy evaluation step. This bias, referred to as \textit{staleness} \citep{bengio2021correcting}, arises when the value function used for bootstrapping at the next state is reflective of an old policy that is currently out of date. Unfortunately, if we want to avoid staleness bias during learning, the length of policy evaluation for each policy must once again depend strongly on the mixing time:
\vspace{-1mm}
\begin{corollary}  \label{corollary:boot}
A bootstrapping algorithm 
with policy $\pi$ in scalable MDP $\mathbb{C}_\sigma$ has a maximum frequency of unbiased policy improvement steps $f^*(\pi) \rightarrow 0$ as $|\mathcal{S}| \rightarrow \infty$.
\end{corollary}
\vspace{-1mm}
\textbf{Bootstrapping for Improvement:} is the process of using bootstrapping for policy improvement as popularized by dynamic programming algorithms such as policy iteration and value iteration, as well as by RL frameworks such as temporal difference (TD) learning and actor-critic. The theoretical foundation of these approaches is the \textit{policy improvement theorem}, which demonstrates the value of taking a greedy action with respect to the estimated action-value function $Q^\pi(s,a)$ at each step. 
The policy improvement theorem then tells us that policy changes towards $\pi'$ are worthwhile as $V^\pi(s) \leq Q^\pi(s,\pi'(s)) \leq V^{\pi'}(s)$ eventually yielding $\pi^*$ in the limit of many changes \citep{SuttonNewBook}. However, this improvement is unlikely to be efficient when mixing times are high. 
To demonstrate this, we decompose value into the transient and limiting components respectively $V^\pi(s) = V_\textrm{trans}^\pi(s) + V_\textrm{lim}^\pi(s)$. While clearly $V^\pi(s) \leq Q^\pi(s,\pi'(s))$, the functions both follow $\pi$ into the future, so their limiting distribution is the same implying that $V_\textrm{trans}^\pi(s) \leq Q_\textrm{trans}^\pi(s,\pi'(s))$ and $V_\textrm{lim}^\pi(s) = Q_\textrm{lim}^\pi(s,\pi'(s))$. 

\begin{corollary}  \label{corollary:improvement}
Policy improvement steps with bootstrapping based on the Bellman optimality operator guarantee monotonic improvement for $V_\textnormal{trans}^\pi$, but do not for $V_\textnormal{lim}^\pi$.
\end{corollary}
\vspace{-1mm}
See Appendix \ref{app:relatedwork} for details on how these ideas connect to relevant off-policy or offline RL approaches and to the literature on catastrophic forgetting in continual RL. 
\vspace{-1mm}
\section{Empirical Analysis of Mixing Behavior}
\vspace{-1mm}
We have established theoretically that scalable MDPs have polynomial mixing times and that polynomial mixing times present significant optimization difficulties for current approaches to RL. However, we still must demonstrate 1) that scalable MDPs are a useful construct for understanding the scaling process within modern continual RL benchmarks and 2) that large mixing times become a significant practical impediment to performing reliable policy evaluation in these domains. Towards this end, we consider empirical scaling of the mixing time with respect to the number of distinct tasks, $|\mathcal{Z}|$, 
and to the task duration, $\tau$, which controls the bottleneck structure.
For this purpose, we evaluate a set of high-quality pretrained policies. 
In both cases, we report empirical results via the dependence of the $\epsilon$-return mixing time (equation \ref{eq:mixingtimes-ret}) on the relative precision with which the reward rate is estimated. 
Presenting the relationship in this way ensures we isolate the contribution that myopic policy evaluation has on the performance relative to the optimum and that high mixing times are not inflated by spurious sources of distributional mismatch that do not hinder policy evaluation.  


\textbf{Empirical Mixing Time Estimation.} In order to quantitatively estimate the $\epsilon$-return mixing time $t_{\textrm{ret}}^{\pi}(\epsilon)$  in equation \ref{eq:mixingtimes-ret}, we need to estimate two terms: the average true reward rate $\rho(\pi)$ (which is agnostic to the start state), and the $h$-step undiscounted return $\rho(h, s_0, \pi)$ from a start state $s_0$ for all $s \in \mathcal{S}$. \textcolor{black}{We can then provide $t_{\textrm{ret}}^{\pi}(\epsilon)$ for any desired value of $\epsilon$.} 
To calculate $\rho(\pi)$ we unroll the policy $\pi$ in the environment for a large number of time steps (more than a million steps) while accumulating the rewards that we received and finally compute the average over the total number of environment interactions. Estimating $\rho(h, s, \pi)$ for every start state $s \in \mathcal{S}$ in equation \ref{eq:mixingtimes-ret} is challenging because of the large state spaces in Atari and Mujoco. Hence we rely on approximations where we choose a subset of states from $\mathcal{S}$ to estimate $\rho(h, s, \pi)$. However, since the estimates could be biased based on the start states we choose, instead of randomly sampling a fixed set of start states, we leverage reservoir sampling \cite{RS} to ensure that the limited fraction of possible start states that we consider is unbiased according to the on-policy distribution. Conservative definitions of the mixing time take  the maximum over start states \citep{kearns2002near}. Here, since we are interested in the characteristic mixing time size, we report the average among start states, which ensures that the mixing times reported are representative of those actually needed for evaluation over the course of on-policy training. Finally, in order to provide an intuitive interpretation for $\epsilon$ across domains, we leverage the \textit{relative error}, $\epsilon/\rho(\pi)$, as a common reference point for comparison. We can then calculate the $\epsilon$-return mixing times for a fixed relative error. 
See Appendix \ref{sec:algo-mixing} for further details\footnote{Our code is available at \href{https://github.com/SharathRaparthy/polynomial_mixing_times.git}{https://github.com/SharathRaparthy/polynomial\_mixing\_times.git}.}. 


\textbf{Setting of Interest.} We focus our experiments on the following scalable MDP formulation that is broadly representative of the majority of work on continual RL \citep{rusu2016progressive,kirkpatrick2017ewc,schwarz2018progress,rolnick2019experience,MER}: 

\begin{example} \label{example1}
(Continual Learning with Passive Task Switching): Consider an environment with tasks $z \in \mathcal{Z}$ and within-task states $x \in \mathcal{X}_z$. The residence time of the region encompassed by task $z$ before moving to another region is fixed for all tasks at $\tau$ for any $\pi$ such that $t^\pi_{\mathcal{X}_z} = \tau \;\; \forall z \in \mathcal{Z}$. Regardless of the way that tasks are connected, the diameter of such an MDP must scale as $D^* \in \Omega(\tau|\mathcal{Z}|)$ because the residence time bounds the minimum possible time to travel between two states in different tasks. \textcolor{black}{Utilizing our formulation of scalable MDPs with initial MDP parameters $\bm{q}_0=(\tau,|\mathcal{Z}|)$, the space of possible proportional scaling functions on $\tau$ only, $|\mathcal{Z}|$ only, and both $\tau$ and $|\mathcal{Z}|$ simultaneously are $\sigma_\tau(\bm{q}_0,\nu)=(\tau+a\nu,|\mathcal{Z}|)$, $\sigma_{|\mathcal{Z}|}(\bm{q}_0,\nu)=(\tau,|\mathcal{Z}|+b\nu)$, and $\sigma_{\tau,|\mathcal{Z}|}(\bm{q}_0,\nu)=(\tau+a\nu,|\mathcal{Z}|+b\nu)$, respectively, where $a,b\in\mathbb{R}$ span the respective spaces.} 
\end{example}

\subsection{A Simple Motivating Example}
We consider a simple $10 \times 10 \times 10$ 3-dimensional grid world environment with 6 actions corresponding to up and down actions in each dimension. In this environment, we can define 1,000 different tasks corresponding to considering each unique state as a goal location. 
At each time-step the agent is sent a 60-dimensional concatenation of six 10-dimensional one-hot vectors. The first three one-hot vectors correspond to the agent's x,y, and z coordinates and the second three correspond to the goal location.

\textbf{Myopic Optimization Bias:} As we have motivated in Section \ref{sec:taskdefinition}, this environment is made to demonstrate the difficulty associated with myopic optimization bias even when task boundaries and their relation are fully observable. We consider a continual episodic RL formulation following the description in Example \ref{example1} where task sequences arise from changing the goal location every $\tau$ steps and the set of goal locations defines the set of tasks, $\mathcal{Z}$. As such, the diameter of the MDP clearly increases as $\tau$ and the number of tasks $|\mathcal{Z}|$ are scaled up, leading to at least linear scaling. As a result, an agent that updates only based on the current episode experiences myopic bias with respect to the steady-state distribution over which the tasks are uniformly balanced.

\textbf{Training Procedure:} The 60-dimensional sparse observation vector is processed by a multi-layer perceptron with two 100 unit hidden layers and Relu activations. The agent performs optimization following episodic REINFORCE. We explore different configurations of this scalable MDP by varying $|\mathcal{Z}|$  from 1 to 1,000 and varying $\tau$ from 100 to 10,000. An important baseline to consider is the extreme of switching tasks after every episode as in multi-task RL (which is generally considered an upper bound on continual RL performance). The agent receives a reward of 1 every step it makes progress towards its goal and a reward of 0 otherwise. This implies that the reward rate of $\pi^*$ is $\rho(\pi^*)=1$ and that the reward rate of a uniform action policy is 0.5 for every task $z \in \mathcal{Z}$ regardless of the distance travelled from the start state for each episode. For each $|\mathcal{Z}|$ and $\tau$, we train the agent for 10,000 total steps and report the average reward rate after training as an average across 300 seeds, which correspond to randomly selected task progressions. See Appendix \ref{app:expdetails} for further details. 
\begin{figure}
    \centering
    \includegraphics[width=0.6\textwidth]{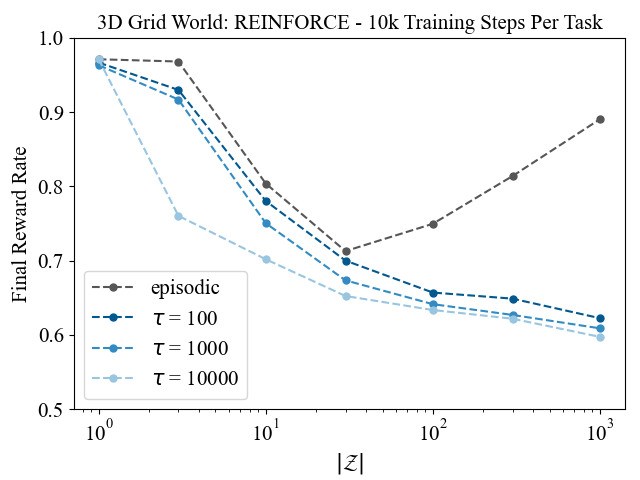}
    \vspace{-1mm}
    \caption{Average reward rate across tasks at the end of training as a function of $|\mathcal{Z}|$ and $\tau$.}
    \label{fig:simple_experiments}
    \vspace{-5mm}
\end{figure}

\textbf{Results:} We display the results of these experiments in Figure \ref{fig:simple_experiments}. It is interesting to notice the episodic transition performance, which can near flawlessly learn a single task in 10,000 steps but struggles significantly with interference in the multitask setting. Performance goes down from 1 to 30 tasks and appears to rebound afterward, which is logical because the similarity between incoming tasks and old tasks is increasing and the total number of training steps across tasks is going up. When both $\tau$  and $|\mathcal{Z}|$ are high at the same time, learning performance is quite poor (not much better than a uniform policy). The starting location is always in the same corner, so some commonalities can be exploited even by a policy experiencing significant interference. Performance bottoms out after fewer tasks when $\tau$ is larger, which makes sense as that is when the myopic bias of optimization is greatest.

\subsection{\textcolor{black}{Overview of Empirical Findings on Atari and Mujoco}}

\textbf{Domains of Focus.} 
We perform experiments involving sequential interaction across 7 Atari environments: \textit{Breakout, Pong, SpaceInvaders, BeamRider, Enduro, SeaQuest} and \textit{Qbert}. This is a typical number of tasks explored in the continual RL literature \citep{rusu2016progressive,kirkpatrick2017ewc,schwarz2018progress,rolnick2019experience,MER}. 
We consider a sequential Atari setup where the individual environments can be seen as sub-regions of a larger environment that are stitched together (see Figure \ref{fig:cont_rl} and Appendix \ref{app:relatedwork} for further details). To demonstrate the validity of our results on vector-valued state spaces, we also consider sequential Mujoco experiments with the following 5 environments: \textit{HalfCheetah, Hopper, Walker2d, Ant} and  \textit{Swimmer}. For both our sequential Atari and Mujoco experiments, we leverage high-performing pretrained policies that are publicly available \cite{rl-zoo3} for mixing time calculations. 
We leverage task labels to use the pretrained model specific to each task as our behavior policy as appropriate when tasks change. 

\begin{figure}
    \centering
    \includegraphics[scale=0.2]{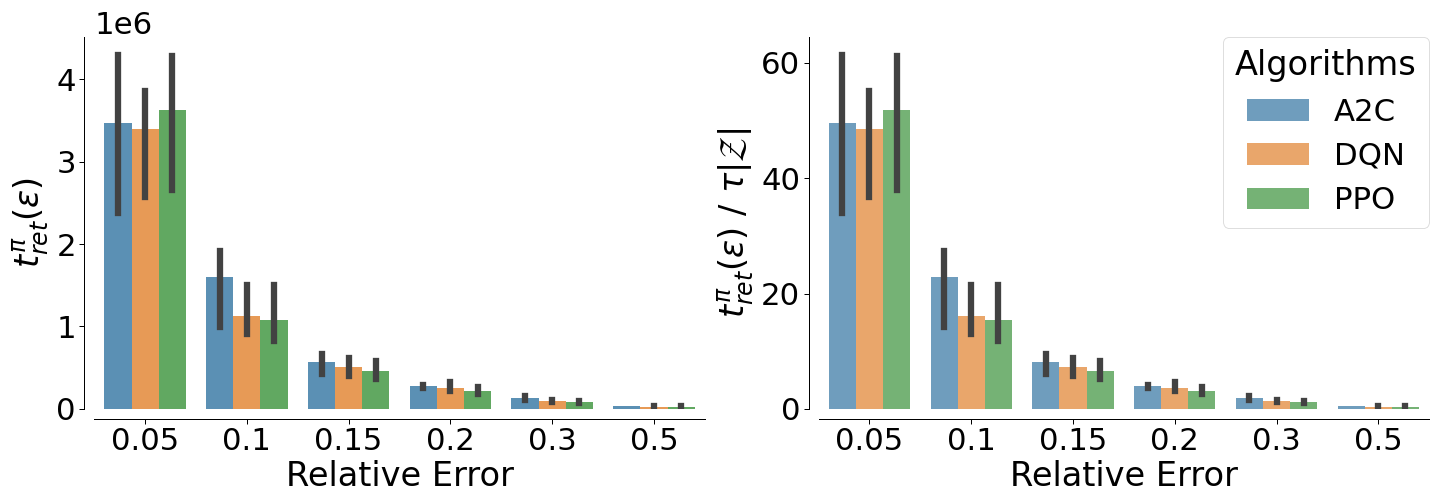}
    \vspace{-2mm}
    \caption{\textbf{Mixing time as a function of relative error for fixed scaling parameters.} \textit{Left}: $\epsilon$-return mixing time (equation \ref{eq:mixingtimes-ret}) as a function of relative error, $\epsilon/\rho(\pi)$, in reward rate estimation for 3 standard algorithms. \textit{Right}: Same as left, here normalized by $\tau|\mathcal{Z}|$ where $\tau=$ 10,000 and $|\mathcal{Z}|=7$. Note the difference in range on the y-axis, which reflects this normalization.}
    \label{fig:precision}
\end{figure}

\textbf{Mixing Time as a Function of Relative Error.} 
As the task connection structure following Example \ref{example1} can take any arbitrary form, we have focused our experiments on random transitions between tasks. 
In Figure \ref{fig:precision} we plot both $t_{\textrm{ret}}^{\pi}(\epsilon)$ and $t_{\textrm{ret}}^{\pi}(\epsilon)/\tau|\mathcal{Z}|$ 
for $\tau=$ 10,000 and $|\mathcal{Z}|=7$ as a function of the relative error of $\rho(\pi, s, h)$ with respect to $\rho(\pi)$\textcolor{black}{, i.e. $\epsilon=\rho(\pi, s, h)-\rho(\pi)$}. 
We see \textcolor{black}{the intuitive fact that higher demands for precision define longer mixing times and} that it is clear that the lower bound on the diameter is a very conservative estimate of the mixing time where we only see  $t_{\textrm{ret}}^{\pi}(\epsilon)/\tau|\mathcal{Z}| < 1$ for reward rates estimated to a poor tolerance of more than 30\% error.

\begin{figure}
    \centering
    \vspace{-4mm}
    \includegraphics[scale=0.2]{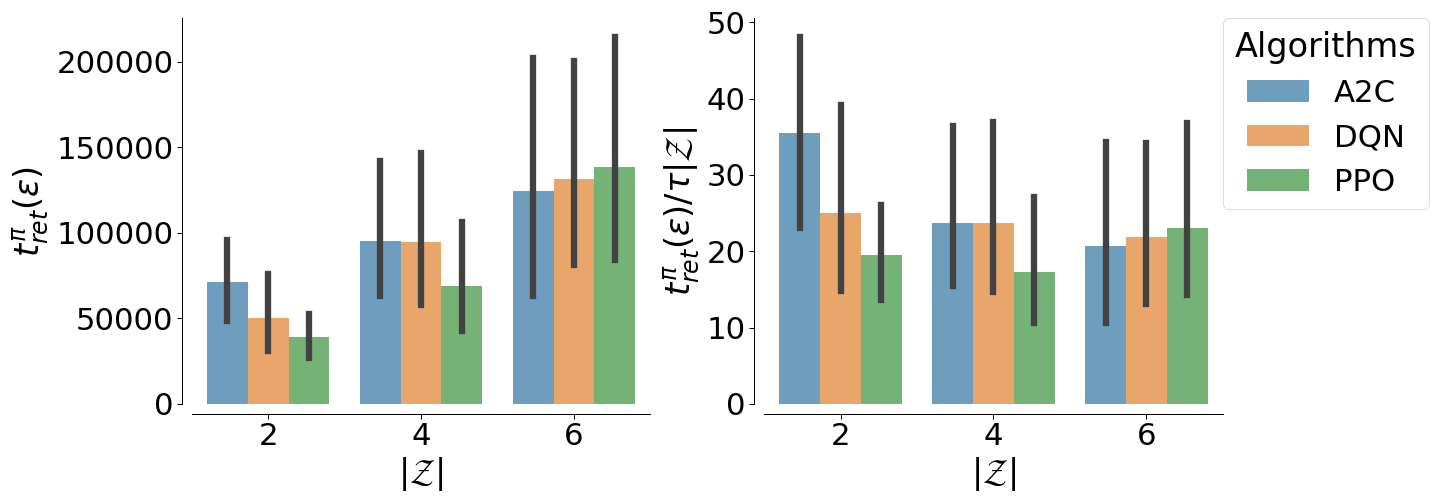}
    \vspace{-2mm}
    \caption{\textbf{Mixing time scaling with the number of tasks $|\mathcal{Z}|$}. \textit{Left}: Average $\epsilon$-return mixing time across different task combinations $\mathcal{Z}$ for different algorithms. \textit{Right}: Same as left, here normalized by $\tau |\mathcal{Z}|$  where $\tau=$1,000. Note the difference in range on the y-axis, which reflects this normalization.}
    \label{fig:tasks}
    \vspace{-5mm}
\end{figure}

\textbf{Scaling $|\mathcal{Z}|$.} We now analyze Example \ref{example1} in the case where the state space and diameter grow as we increase the number of tasks $|\mathcal{Z}|$. Specifically we scale $|\mathcal{Z}|$ from 2 to 4 to 6 as $\tau$ is kept fixed at 1,000 and generate 10 random task combinations from the list of 7 for each value of $|\mathcal{Z}|$. In Figure \ref{fig:tasks} we plot the resulting mixing times for a precision of 10\% error in approximating the reward rate. We find that there is a linear trend in the expected mixing time and not just the theoretical lower bound. 



\begin{figure}
     \centering
     \vspace{-7mm}
     \begin{subfigure}[b]{0.5\columnwidth}
         \centering
         \includegraphics[width=1.07\columnwidth]{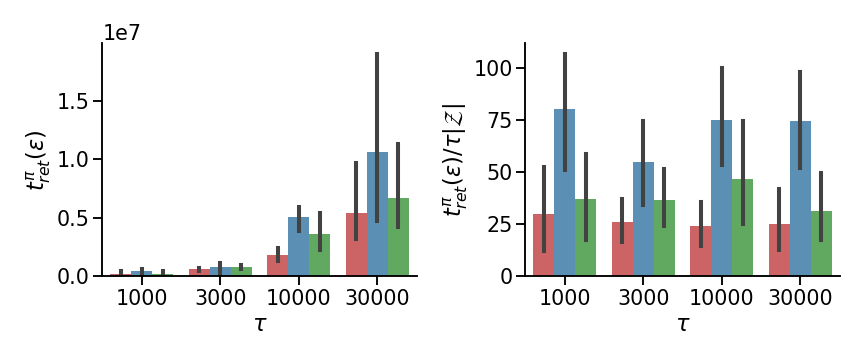}
         \vspace{-7mm}
         \caption{\textcolor{black}{Continual Mujoco $|\mathcal{Z}| = 5$}}
         \label{fig:cont-mujoco}
     \end{subfigure}
     \hfill
     \begin{subfigure}[b]{0.49\columnwidth}
         \centering
         \includegraphics[width=1.1\columnwidth]{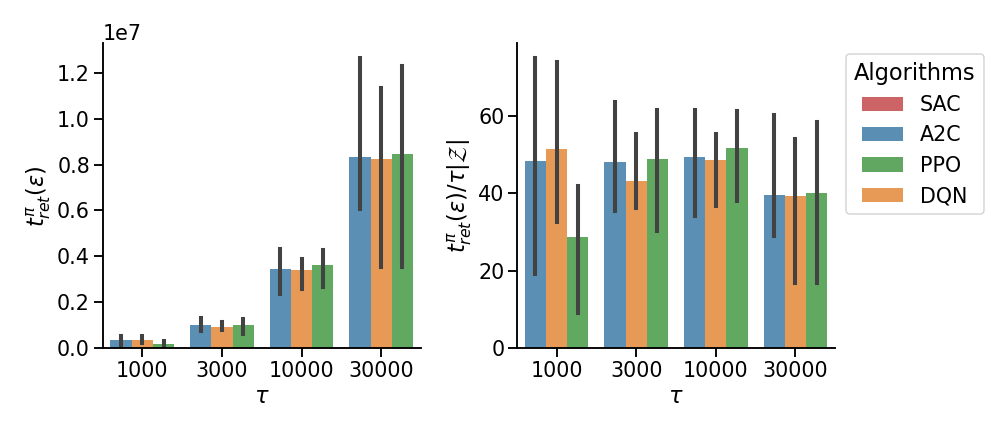}
         \vspace{-7mm}
         \caption{\textcolor{black}{Continual Atari $|\mathcal{Z}| = 7$}}
         \label{fig:cont-atari}
     \end{subfigure}
     \vspace{-5mm}
        \caption{\textcolor{black}{\textbf{Mixing time scaling with the task duration, $\tau$}. Left (a, b): $\epsilon$-return mixing time across different $\tau$ values for different algorithms. Right (a, b): Same as left, here normalized by $\tau|\mathcal{Z}|$.  Note the difference in range on the y-axis and the distinct subset of algorithms tested (see legend).}}
        \label{fig:switching}
\end{figure}

\textbf{Scaling $\tau$.} We finally analyze Example \ref{example1} in the case where the state space is held constant but the diameter grows with the increasingly severe bottleneck structure as we increase $\tau$. Here we consider $|\mathcal{Z}|=7$ and $\tau=$ 1,000, 3,000, 10,000, and 30,000, while calculating $\rho(h, s, \pi)$ and $\rho(\pi)$ by unrolling the policy $\pi$ for $\tau \times $1,000 environment steps each. In Figure \ref{fig:switching} we once again plot both $t_{\textrm{ret}}^{\pi}(\epsilon)$ and the value normalized by the diameter lower bound $t_{\textrm{ret}}^{\pi}(\epsilon)/\tau|\mathcal{Z}|$ to find that empirical average mixing times scale linearly with increasing $\tau$ and not just the lower bound. 
\vspace{-1mm}
\subsection{\textcolor{black}{Implications For Continual RL}}
\vspace{-1mm}
In summary, we have observed in Figure \ref{fig:precision} that average mixing times reach a staggering high for lower relative errors, and shorter mixing times are associated with very high relative errors. Additionally, in Figures \ref{fig:tasks} and \ref{fig:switching} we observed that, for a broad class of algorithms, there seems to be a linear dependency of the expected mixing time with increasing $|\mathcal{Z}|$ and increasing $\tau$.

\textbf{Optimization Instability for Continual Atari.} While we have empirically established that mixing times scale to be quite large in the Atari domain following Example \ref{example1} and established theoretically how RL approaches struggle when this is the case in Section \ref{sec:myopicbias}, we would like to emphasize that the prior literature applying deep RL to continual learning on Atari has empirically come to the same conclusion. Indeed, \citep{kirkpatrick2017ewc} looked at continual learning across 10 Atari games with a randomized duration between task switches and found that all approaches were significantly outperformed by the single task baseline, reflecting significant instability in optimizing across tasks. Additionally, perhaps the most common setting in the literature considered by \citep{schwarz2018progress}, \citep{rolnick2019experience}, and \citep{gitrepo} is to train on a set of tasks where $|\mathcal{Z}|=6$ and $\tau=$ 50,000,000. Note that extrapolating from our results would indicate average mixing times on the order of $\approx 15$ billion steps for a precision of 5\% error or $\approx 5$ billion steps for a precision of 10\% error. Across a number of RL approaches tried on these domains, as we would expect, all approaches have experienced significant instability in their learning curves. A very popular failure mode is approaches that display too much plasticity such that their performance for a task goes through big peaks and valleys resulting from myopic bias towards optimization on the current task. This phenomena is often known in the literature as catastrophic forgetting \citep{CF}.

\vspace{-1mm}
\section{\textcolor{black}{Inspiring New Approaches for Continual RL}}
\vspace{-1mm}
While the focus of our paper is on highlighting the difficulty of the polynomial mixing time problem, not proposing an approach to solve it, we believe that our analysis provides significant insights that the community will be able to draw on to build better 
approaches for continual RL: 

\textcolor{black}{\textbf{Direct Steady-State Estimation:}} In the context of commonly considered problems that follow the high-level structure of Example \ref{example1}, we have showcased the deep connection between catastrophic forgetting and myopic bias in the presence of very large mixing times. In Appendix \ref{app:relatedwork} we demonstrate how this perspective explains some of the value of replay based approaches in this setting. Moreover, this perspective naturally points towards approaches that directly reason over the steady-state distribution as a natural solution to the catastrophic forgetting problem. In Appendix \ref{app:expdetails} we consider simple tabular RL experiments across 3 example classes of scalable MDPs. Our analysis demonstrates that on-policy and off-policy versions of Q-Learning that perform policy evaluation using an estimate of the steady-state distribution, derived from matrix inversion of a tabular environment model, consistently outperform standard model-based and model-free baselines in terms of lifelong regret as environments are scaled up. Even though estimating the steady-state distribution would clearly be quite challenging in complex state and observation spaces, recent approaches have made great strides towards developing practical and scalable approaches leveraging data buffers \citep{vpm}.

\textcolor{black}{\textbf{Tracking Approaches:} It was originally noted by \citep{tracking} that it could be a better strategy even in stationary environments with temporal coherence to learn to track the best local solution instead of attempting to converge to a globally optimal fixed policy. This indeed, may be a necessity in environments that experience extreme temporal coherence governed by high mixing times and could provide a formalization of the benefit for approaches based on meta-learning in this context as noted by \citep{tracking}. For example, \citep{kim2022influencing} recently was successful in addressing the non-stationarity of multi-agent RL by learning to converge to a recurrent set of joint policies rather than a single joint policy.}

\textbf{Looking Forward.}
In this work, we have considered the implicit premise of the current continual RL literature that results from currently manageable small-scale experiments are indicative of performance for the large-scale aspirational use cases of the future. In particular, we have highlighted that mixing times will scale significantly as the problems we deal with are scaled and how traditional approaches to RL are ill-suited to deal with this. 
We hope that our work will encourage the community to carefully consider the theoretical implications of scaling environments, especially with regard to how proposed continual RL algorithms scale with the mixing time.

\section*{Acknowledgements}
We would like to thank Riashat Islam, Gerald Tesauro, Miao Liu and Sarthak Mittal for helpful discussions. Irina Rish acknowledges the support from Canada CIFAR AI Chair
Program and from the Canada Excellence Research Chairs
Program. We thank the IBM Cognitive Compute Cluster, Mila cluster and Compute Canada for providing computational resources.
\bibliographystyle{unsrtnat}
\bibliography{neurips_2022}

\begin{thebibliography}{64}
\providecommand{\natexlab}[1]{#1}
\providecommand{\url}[1]{\texttt{#1}}
\expandafter\ifx\csname urlstyle\endcsname\relax
  \providecommand{\doi}[1]{doi: #1}\else
  \providecommand{\doi}{doi: \begingroup \urlstyle{rm}\Url}\fi

\bibitem[Khetarpal et~al.(2020)Khetarpal, Riemer, Rish, and Precup]{crlsurvey}
Khimya Khetarpal, Matthew Riemer, Irina Rish, and Doina Precup.
\newblock Towards continual reinforcement learning: A review and perspectives.
\newblock \emph{arXiv preprint arXiv:2012.13490}, 2020.

\bibitem[Schwartz(1993)]{rlearning}
Anton Schwartz.
\newblock A reinforcement learning method for maximizing undiscounted rewards.
\newblock In \emph{Proceedings of the tenth international conference on machine
  learning}, volume 298, pages 298--305, 1993.

\bibitem[Naik et~al.(2019)Naik, Shariff, Yasui, Yao, and
  Sutton]{stopdiscounting}
Abhishek Naik, Roshan Shariff, Niko Yasui, Hengshuai Yao, and Richard~S Sutton.
\newblock Discounted reinforcement learning is not an optimization problem.
\newblock \emph{arXiv preprint arXiv:1910.02140}, 2019.

\bibitem[Nota and Thomas(2019)]{notagradients}
Chris Nota and Philip~S. Thomas.
\newblock Is the policy gradient a gradient?
\newblock \emph{CoRR}, abs/1906.07073, 2019.
\newblock URL \url{http://arxiv.org/abs/1906.07073}.

\bibitem[Sutton and Barto(2018)]{SuttonNewBook}
Richard~S Sutton and Andrew~G Barto.
\newblock \emph{Reinforcement learning: An introduction}.
\newblock 2018.

\bibitem[Puterman(1994)]{puterman1994markov}
ML~Puterman.
\newblock Markov decision processes. 1994.
\newblock \emph{Jhon Wiley \& Sons, New Jersey}, 1994.

\bibitem[Wan et~al.(2020)Wan, Naik, and Sutton]{learningandplanning}
Yi~Wan, Abhishek Naik, and Richard~S Sutton.
\newblock Learning and planning in average-reward markov decision processes.
\newblock \emph{arXiv preprint arXiv:2006.16318}, 2020.

\bibitem[Bertsekas(1998)]{bertsekas98}
Dimitri~P Bertsekas.
\newblock A new value iteration method for the average cost dynamic programming
  problem.
\newblock \emph{SIAM journal on control and optimization}, 36\penalty0
  (2):\penalty0 742--759, 1998.

\bibitem[Kearns and Singh(2002)]{kearns2002near}
Michael Kearns and Satinder Singh.
\newblock Near-optimal reinforcement learning in polynomial time.
\newblock \emph{Machine learning}, 49\penalty0 (2):\penalty0 209--232, 2002.

\bibitem[Levin and Peres(2017)]{mixingtimesbook}
David~A Levin and Yuval Peres.
\newblock \emph{Markov chains and mixing times}, volume 107.
\newblock American Mathematical Soc., 2017.

\bibitem[Jaksch et~al.(2010)Jaksch, Ortner, and Auer]{jaksch2010near}
Thomas Jaksch, Ronald Ortner, and Peter Auer.
\newblock Near-optimal regret bounds for reinforcement learning.
\newblock \emph{Journal of Machine Learning Research}, 11\penalty0 (4), 2010.

\bibitem[De~Lange et~al.(2021)De~Lange, Aljundi, Masana, Parisot, Jia,
  Leonardis, Slabaugh, and Tuytelaars]{delange}
Matthias De~Lange, Rahaf Aljundi, Marc Masana, Sarah Parisot, Xu~Jia,
  Ale{\v{s}} Leonardis, Gregory Slabaugh, and Tinne Tuytelaars.
\newblock A continual learning survey: Defying forgetting in classification
  tasks.
\newblock \emph{IEEE transactions on pattern analysis and machine
  intelligence}, 44\penalty0 (7):\penalty0 3366--3385, 2021.

\bibitem[Zahavy et~al.(2020)Zahavy, Cohen, Kaplan, and
  Mansour]{zahavy2020unknown}
Tom Zahavy, Alon Cohen, Haim Kaplan, and Yishay Mansour.
\newblock Unknown mixing times in apprenticeship and reinforcement learning.
\newblock In \emph{Conference on Uncertainty in Artificial Intelligence}, pages
  430--439. PMLR, 2020.

\bibitem[Bengio et~al.(2021)Bengio, Pineau, and Precup]{bengio2021correcting}
Emmanuel Bengio, Joelle Pineau, and Doina Precup.
\newblock Correcting momentum in temporal difference learning.
\newblock \emph{arXiv preprint arXiv:2106.03955}, 2021.

\bibitem[Vitter(1985)]{RS}
Jeffrey~S Vitter.
\newblock Random sampling with a reservoir.
\newblock \emph{ACM Transactions on Mathematical Software (TOMS)}, 11\penalty0
  (1):\penalty0 37--57, 1985.

\bibitem[Rusu et~al.(2016)Rusu, Rabinowitz, Desjardins, Soyer, Kirkpatrick,
  Kavukcuoglu, Pascanu, and Hadsell]{rusu2016progressive}
Andrei~A Rusu, Neil~C Rabinowitz, Guillaume Desjardins, Hubert Soyer, James
  Kirkpatrick, Koray Kavukcuoglu, Razvan Pascanu, and Raia Hadsell.
\newblock Progressive neural networks.
\newblock \emph{arXiv preprint arXiv:1606.04671}, 2016.

\bibitem[Kirkpatrick et~al.(2017)Kirkpatrick, Pascanu, Rabinowitz, Veness,
  Desjardins, Rusu, Milan, Quan, Ramalho, Grabska-Barwinska,
  et~al.]{kirkpatrick2017ewc}
James Kirkpatrick, Razvan Pascanu, Neil Rabinowitz, Joel Veness, Guillaume
  Desjardins, Andrei~A Rusu, Kieran Milan, John Quan, Tiago Ramalho, Agnieszka
  Grabska-Barwinska, et~al.
\newblock Overcoming catastrophic forgetting in neural networks.
\newblock \emph{Proceedings of the National Academy of Sciences of the United
  States of America}, 114\penalty0 (13):\penalty0 3521--3526, 2017.

\bibitem[Schwarz et~al.(2018)Schwarz, Czarnecki, Luketina, Grabska-Barwinska,
  Teh, Pascanu, and Hadsell]{schwarz2018progress}
Jonathan Schwarz, Wojciech Czarnecki, Jelena Luketina, Agnieszka
  Grabska-Barwinska, Yee~Whye Teh, Razvan Pascanu, and Raia Hadsell.
\newblock Progress \& compress: A scalable framework for continual learning.
\newblock In \emph{International Conference on Machine Learning}, pages
  4528--4537, 2018.

\bibitem[Rolnick et~al.(2019)Rolnick, Ahuja, Schwarz, Lillicrap, and
  Wayne]{rolnick2019experience}
David Rolnick, Arun Ahuja, Jonathan Schwarz, Timothy Lillicrap, and Gregory
  Wayne.
\newblock Experience replay for continual learning.
\newblock \emph{Advances in Neural Information Processing Systems},
  32:\penalty0 350--360, 2019.

\bibitem[Riemer et~al.(2018)Riemer, Cases, Ajemian, Liu, Rish, Tu, and
  Tesauro]{MER}
Matthew Riemer, Ignacio Cases, Robert Ajemian, Miao Liu, Irina Rish, Yuhai Tu,
  and Gerald Tesauro.
\newblock Learning to learn without forgetting by maximizing transfer and
  minimizing interference.
\newblock \emph{arXiv preprint arXiv:1810.11910}, 2018.

\bibitem[Raffin(2020)]{rl-zoo3}
Antonin Raffin.
\newblock Rl baselines3 zoo.
\newblock \url{https://github.com/DLR-RM/rl-baselines3-zoo}, 2020.

\bibitem[AGI-Labs(2021)]{gitrepo}
AGI-Labs.
\newblock Continual rl project.
\newblock \url{https://github.com/AGI-Labs/continual_rl}, 2021.

\bibitem[McCloskey and Cohen(1989{\natexlab{a}})]{CF}
Michael McCloskey and Neal~J Cohen.
\newblock Catastrophic interference in connectionist networks: The sequential
  learning problem.
\newblock \emph{Psychology of learning and motivation}, 24:\penalty0 109--165,
  1989{\natexlab{a}}.

\bibitem[Wen et~al.(2020{\natexlab{a}})Wen, Dai, Li, and Schuurmans]{vpm}
Junfeng Wen, Bo~Dai, Lihong Li, and Dale Schuurmans.
\newblock Batch stationary distribution estimation.
\newblock \emph{arXiv preprint arXiv:2003.00722}, 2020{\natexlab{a}}.

\bibitem[Sutton et~al.(2007)Sutton, Koop, and Silver]{tracking}
Richard~S Sutton, Anna Koop, and David Silver.
\newblock On the role of tracking in stationary environments.
\newblock In \emph{Proceedings of the 24th international conference on Machine
  learning}, pages 871--878, 2007.

\bibitem[Kim et~al.(2022)Kim, Riemer, Liu, Foerster, Everett, Sun, Tesauro, and
  How]{kim2022influencing}
Dong-Ki Kim, Matthew Riemer, Miao Liu, Jakob~N Foerster, Michael Everett,
  Chuangchuang Sun, Gerald Tesauro, and Jonathan~P How.
\newblock Influencing long-term behavior in multiagent reinforcement learning.
\newblock \emph{arXiv preprint arXiv:2203.03535}, 2022.

\bibitem[McCloskey and Cohen(1989{\natexlab{b}})]{mccloskey1989catastrophic}
Michael McCloskey and Neal~J Cohen.
\newblock Catastrophic interference in connectionist networks: The sequential
  learning problem.
\newblock \emph{Psychology of learning and motivation, 24:109–165},
  1989{\natexlab{b}}.

\bibitem[Carpenter and Grossberg(1987)]{dilemma}
Gail~A Carpenter and Stephen Grossberg.
\newblock A massively parallel architecture for a self-organizing neural
  pattern recognition machine.
\newblock \emph{Computer vision, graphics, and image processing}, 37\penalty0
  (1):\penalty0 54--115, 1987.

\bibitem[Lin(1992)]{lin1992self}
Long-Ji Lin.
\newblock Self-improving reactive agents based on reinforcement learning,
  planning and teaching.
\newblock \emph{Machine learning}, 8\penalty0 (3-4):\penalty0 293--321, 1992.

\bibitem[Isele and Cosgun(2018)]{isele2018selective}
David Isele and Akansel Cosgun.
\newblock Selective experience replay for lifelong learning.
\newblock In \emph{Proceedings of the AAAI Conference on Artificial
  Intelligence}, volume~32, 2018.

\bibitem[Riemer et~al.(2019)Riemer, Klinger, Bouneffouf, and
  Franceschini]{riemer2019scalable}
Matthew Riemer, Tim Klinger, Djallel Bouneffouf, and Michele Franceschini.
\newblock Scalable recollections for continual lifelong learning.
\newblock In \emph{Proceedings of the AAAI Conference on Artificial
  Intelligence}, volume~33, pages 1352--1359, 2019.

\bibitem[Pan et~al.(2018)Pan, Zaheer, White, Patterson, and
  White]{pan2018organizing}
Yangchen Pan, Muhammad Zaheer, Adam White, Andrew Patterson, and Martha White.
\newblock Organizing experience: a deeper look at replay mechanisms for
  sample-based planning in continuous state domains.
\newblock \emph{arXiv preprint arXiv:1806.04624}, 2018.

\bibitem[Duff(2002)]{duff2002optimal}
Michael~O'Gordon Duff.
\newblock \emph{Optimal Learning: Computational procedures for Bayes-adaptive
  Markov decision processes}.
\newblock PhD thesis, University of Massachusetts at Amherst, 2002.

\bibitem[Ross et~al.(2007)Ross, Chaib-draa, and Pineau]{BAMDP}
Stephane Ross, Brahim Chaib-draa, and Joelle Pineau.
\newblock Bayes-adaptive pomdps.
\newblock In \emph{NIPS}, pages 1225--1232, 2007.

\bibitem[Zintgraf et~al.(2019)Zintgraf, Shiarlis, Igl, Schulze, Gal, Hofmann,
  and Whiteson]{varibad}
Luisa Zintgraf, Kyriacos Shiarlis, Maximilian Igl, Sebastian Schulze, Yarin
  Gal, Katja Hofmann, and Shimon Whiteson.
\newblock Varibad: A very good method for bayes-adaptive deep rl via
  meta-learning.
\newblock \emph{arXiv preprint arXiv:1910.08348}, 2019.

\bibitem[Choi et~al.(2000)Choi, Yeung, and Zhang]{HMMDP}
Samuel~PM Choi, Dit-Yan Yeung, and Nevin~L Zhang.
\newblock Hidden-mode markov decision processes for nonstationary sequential
  decision making.
\newblock In \emph{Sequence Learning}, pages 264--287. Springer, 2000.

\bibitem[Xie et~al.(2020)Xie, Harrison, and Finn]{xie2020deep}
Annie Xie, James Harrison, and Chelsea Finn.
\newblock Deep reinforcement learning amidst lifelong non-stationarity, 2020.

\bibitem[Ong et~al.(2010)Ong, Png, Hsu, and Lee]{MOMDP}
Sylvie~CW Ong, Shao~Wei Png, David Hsu, and Wee~Sun Lee.
\newblock Planning under uncertainty for robotic tasks with mixed
  observability.
\newblock \emph{The International Journal of Robotics Research}, 29\penalty0
  (8):\penalty0 1053--1068, 2010.

\bibitem[Foerster et~al.(2017)Foerster, Chen, Al-Shedivat, Whiteson, Abbeel,
  and Mordatch]{lola}
Jakob~N Foerster, Richard~Y Chen, Maruan Al-Shedivat, Shimon Whiteson, Pieter
  Abbeel, and Igor Mordatch.
\newblock Learning with opponent-learning awareness.
\newblock \emph{arXiv preprint arXiv:1709.04326}, 2017.

\bibitem[Al-Shedivat et~al.(2018)Al-Shedivat, Bansal, Burda, Sutskever,
  Mordatch, and Abbeel]{CMAML}
Maruan Al-Shedivat, Trapit Bansal, Yuri Burda, Ilya Sutskever, Igor Mordatch,
  and Pieter Abbeel.
\newblock Continuous adaptation via meta-learning in nonstationary and
  competitive environments.
\newblock \emph{ICLR}, 2018.

\bibitem[Kim et~al.(2021)Kim, Liu, Riemer, Sun, Abdulhai, Habibi, Lopez-Cot,
  Tesauro, and How]{metamapg}
Dong~Ki Kim, Miao Liu, Matthew~D Riemer, Chuangchuang Sun, Marwa Abdulhai,
  Golnaz Habibi, Sebastian Lopez-Cot, Gerald Tesauro, and Jonathan How.
\newblock A policy gradient algorithm for learning to learn in multiagent
  reinforcement learning.
\newblock In \emph{International Conference on Machine Learning}, pages
  5541--5550. PMLR, 2021.

\bibitem[Kearns and Koller(1999)]{kearns1999efficient}
Michael Kearns and Daphne Koller.
\newblock Efficient reinforcement learning in factored mdps.
\newblock In \emph{IJCAI}, volume~16, pages 740--747, 1999.

\bibitem[Boutilier et~al.(2000)Boutilier, Dearden, and
  Goldszmidt]{boutilier2000stochastic}
Craig Boutilier, Richard Dearden, and Mois{\'e}s Goldszmidt.
\newblock Stochastic dynamic programming with factored representations.
\newblock \emph{Artificial intelligence}, 121\penalty0 (1-2):\penalty0 49--107,
  2000.

\bibitem[Strehl et~al.(2007)Strehl, Diuk, and Littman]{strehl2007efficient}
Alexander~L Strehl, Carlos Diuk, and Michael~L Littman.
\newblock Efficient structure learning in factored-state mdps.
\newblock In \emph{AAAI}, volume~7, pages 645--650, 2007.

\bibitem[Osband and Van~Roy(2014)]{osband2014near}
Ian Osband and Benjamin Van~Roy.
\newblock Near-optimal reinforcement learning in factored mdps.
\newblock \emph{Advances in Neural Information Processing Systems},
  27:\penalty0 604--612, 2014.

\bibitem[Chitnis et~al.(2020)Chitnis, Silver, Kim, Kaelbling, and
  Lozano-Perez]{chitnis2020camps}
Rohan Chitnis, Tom Silver, Beomjoon Kim, Leslie~Pack Kaelbling, and Tomas
  Lozano-Perez.
\newblock Camps: Learning context-specific abstractions for efficient planning
  in factored mdps.
\newblock \emph{arXiv preprint arXiv:2007.13202}, 2020.

\bibitem[Abdulhai et~al.(2021)Abdulhai, Kim, Riemer, Liu, Tesauro, and
  How]{abdulhai2021context}
Marwa Abdulhai, Dong-Ki Kim, Matthew Riemer, Miao Liu, Gerald Tesauro, and
  Jonathan~P How.
\newblock Context-specific representation abstraction for deep option learning.
\newblock \emph{arXiv preprint arXiv:2109.09876}, 2021.

\bibitem[Sutton et~al.(1999)Sutton, Precup, and Singh]{options}
Richard~S Sutton, Doina Precup, and Satinder Singh.
\newblock Between mdps and semi-mdps: A framework for temporal abstraction in
  reinforcement learning.
\newblock \emph{Artificial intelligence}, 112\penalty0 (1-2):\penalty0
  181--211, 1999.

\bibitem[Padakandla et~al.(2019)Padakandla, Bhatnagar,
  et~al.]{nonstationaryrlenvs}
Sindhu Padakandla, Shalabh Bhatnagar, et~al.
\newblock Reinforcement learning in non-stationary environments.
\newblock \emph{arXiv preprint arXiv:1905.03970}, 2019.

\bibitem[Lecarpentier and Rachelson(2019)]{lecarpentier2019non}
Erwan Lecarpentier and Emmanuel Rachelson.
\newblock Non-stationary markov decision processes, a worst-case approach using
  model-based reinforcement learning.
\newblock In \emph{Advances in Neural Information Processing Systems}, pages
  7214--7223, 2019.

\bibitem[Chandak et~al.(2020)Chandak, Theocharous, Shankar, Mahadevan, White,
  and Thomas]{chandak}
Yash Chandak, Georgios Theocharous, Shiv Shankar, Sridhar Mahadevan, Martha
  White, and Philip~S Thomas.
\newblock Optimizing for the future in non-stationary mdps.
\newblock \emph{arXiv preprint arXiv:2005.08158}, 2020.

\bibitem[Sutton(1991)]{sutton1991dyna}
Richard~S Sutton.
\newblock Dyna, an integrated architecture for learning, planning, and
  reacting.
\newblock \emph{ACM Sigart Bulletin}, 2\penalty0 (4):\penalty0 160--163, 1991.

\bibitem[Janner et~al.(2019)Janner, Fu, Zhang, and Levine]{mbpo}
Michael Janner, Justin Fu, Marvin Zhang, and Sergey Levine.
\newblock When to trust your model: Model-based policy optimization.
\newblock \emph{arXiv preprint arXiv:1906.08253}, 2019.

\bibitem[Holland et~al.(2018)Holland, Talvitie, and Bowling]{holland2018effect}
G~Zacharias Holland, Erin~J Talvitie, and Michael Bowling.
\newblock The effect of planning shape on dyna-style planning in
  high-dimensional state spaces.
\newblock \emph{arXiv preprint arXiv:1806.01825}, 2018.

\bibitem[Precup(2000)]{importancesampling}
Doina Precup.
\newblock Eligibility traces for off-policy policy evaluation.
\newblock \emph{Computer Science Department Faculty Publication Series},
  page~80, 2000.

\bibitem[Levine et~al.(2020)Levine, Kumar, Tucker, and Fu]{levine2020offline}
Sergey Levine, Aviral Kumar, George Tucker, and Justin Fu.
\newblock Offline reinforcement learning: Tutorial, review, and perspectives on
  open problems.
\newblock \emph{arXiv preprint arXiv:2005.01643}, 2020.

\bibitem[Hallak and Mannor(2017)]{hallak2017consistent}
Assaf Hallak and Shie Mannor.
\newblock Consistent on-line off-policy evaluation.
\newblock In \emph{International Conference on Machine Learning}, pages
  1372--1383. PMLR, 2017.

\bibitem[Gelada and Bellemare(2019)]{gelada2019off}
Carles Gelada and Marc~G Bellemare.
\newblock Off-policy deep reinforcement learning by bootstrapping the covariate
  shift.
\newblock In \emph{Proceedings of the AAAI Conference on Artificial
  Intelligence}, volume~33, pages 3647--3655, 2019.

\bibitem[Wen et~al.(2020{\natexlab{b}})Wen, Dai, Li, and
  Schuurmans]{wen2020batch}
Junfeng Wen, Bo~Dai, Lihong Li, and Dale Schuurmans.
\newblock Batch stationary distribution estimation.
\newblock \emph{arXiv preprint arXiv:2003.00722}, 2020{\natexlab{b}}.

\bibitem[Nachum et~al.(2019{\natexlab{a}})Nachum, Chow, Dai, and
  Li]{nachum2019dualdice}
Ofir Nachum, Yinlam Chow, Bo~Dai, and Lihong Li.
\newblock Dualdice: Behavior-agnostic estimation of discounted stationary
  distribution corrections.
\newblock \emph{arXiv preprint arXiv:1906.04733}, 2019{\natexlab{a}}.

\bibitem[Nachum et~al.(2019{\natexlab{b}})Nachum, Dai, Kostrikov, Chow, Li, and
  Schuurmans]{nachum2019algaedice}
Ofir Nachum, Bo~Dai, Ilya Kostrikov, Yinlam Chow, Lihong Li, and Dale
  Schuurmans.
\newblock Algaedice: Policy gradient from arbitrary experience.
\newblock \emph{arXiv preprint arXiv:1912.02074}, 2019{\natexlab{b}}.

\bibitem[Tang et~al.(2019)Tang, Feng, Li, Zhou, and Liu]{tang2019doubly}
Ziyang Tang, Yihao Feng, Lihong Li, Dengyong Zhou, and Qiang Liu.
\newblock Doubly robust bias reduction in infinite horizon off-policy
  estimation.
\newblock \emph{arXiv preprint arXiv:1910.07186}, 2019.

\bibitem[Nachum and Dai(2020)]{nachum2020reinforcement}
Ofir Nachum and Bo~Dai.
\newblock Reinforcement learning via fenchel-rockafellar duality.
\newblock \emph{arXiv preprint arXiv:2001.01866}, 2020.

\bibitem[Osband and Van~Roy(2016)]{osband2016lower}
Ian Osband and Benjamin Van~Roy.
\newblock On lower bounds for regret in reinforcement learning.
\newblock \emph{arXiv preprint arXiv:1608.02732}, 2016.

\end{thebibliography}

\section*{Checklist}


\begin{enumerate}

\item For all authors...
\begin{enumerate}
  \item Do the main claims made in the abstract and introduction accurately reflect the paper's contributions and scope?
    \answerYes{}
  \item Did you describe the limitations of your work?
    \answerYes{}
  \item Did you discuss any potential negative societal impacts of your work?
    \answerYes{}
  \item Have you read the ethics review guidelines and ensured that your paper conforms to them?
    \answerYes{}
\end{enumerate}

\item If you are including theoretical results...
\begin{enumerate}
  \item Did you state the full set of assumptions of all theoretical results?
    \answerYes{}
        \item Did you include complete proofs of all theoretical results?
    \answerYes{}
\end{enumerate}

\item If you ran experiments...
\begin{enumerate}
  \item Did you include the code, data, and instructions needed to reproduce the main experimental results (either in the supplemental material or as a URL)?
    \answerYes{}
  \item Did you specify all the training details (e.g., data splits, hyperparameters, how they were chosen)?
    \answerYes{}
        \item Did you report error bars (e.g., with respect to the random seed after running experiments multiple times)?
    \answerYes{}
        \item Did you include the total amount of compute and the type of resources used (e.g., type of GPUs, internal cluster, or cloud provider)?
    \answerYes{}
\end{enumerate}

\item If you are using existing assets (e.g., code, data, models) or curating/releasing new assets...
\begin{enumerate}
  \item If your work uses existing assets, did you cite the creators?
    \answerYes{}
  \item Did you mention the license of the assets?
    \answerNA{}
  \item Did you include any new assets either in the supplemental material or as a URL?
    \answerNA{}
  \item Did you discuss whether and how consent was obtained from people whose data you're using/curating?
    \answerNA{}
  \item Did you discuss whether the data you are using/curating contains personally identifiable information or offensive content?
    \answerNA{}
\end{enumerate}

\item If you used crowdsourcing or conducted research with human subjects...
\begin{enumerate}
  \item Did you include the full text of instructions given to participants and screenshots, if applicable?
    \answerNA{}
  \item Did you describe any potential participant risks, with links to Institutional Review Board (IRB) approvals, if applicable?
    \answerNA{}
  \item Did you include the estimated hourly wage paid to participants and the total amount spent on participant compensation?
    \answerNA{}
\end{enumerate}

\end{enumerate}
\appendix

\appendix

\section{Connections To Literature On Continual And Off-Policy RL} \label{app:relatedwork}



In this section, we begin by highlighting connections to the literature on catastrophic forgetting and task structure in continual RL.
Moreover, in the main text we primarily discussed the connection between polynomial mixing times and implications for on-policy model-free and on-policy model-based RL approaches. Here we will discuss connections to off-policy approaches as well.   

\subsection{Connection to Literature on Continual RL}

\textbf{Catastrophic Forgetting and Experience Replay:} In the main text, we began to highlight connections between myopic optimization bias and catastrophic forgetting \citep{mccloskey1989catastrophic} in continual RL. Indeed, this often results from myopic updates that do not properly consider the long-term distribution of experiences needed to properly address the stability-plasticity dilemma \citep{dilemma} of continual RL as explained in \citep{crlsurvey}. A particularly successful approach in this regard has been those based on experience replay \citep{lin1992self} such as recent approaches to continual RL \citep{isele2018selective,MER,rolnick2019experience} or approaches that use a scalable generative form of replay \citep{riemer2019scalable}. 
As highlighted by \citep{pan2018organizing}, experience replay is closely related to model-based RL approaches. As such, we should only expect experience replay to help us optimize for the infinite horizon objective to the extent that our replay buffer reflects the stationary distribution of the current policy. The likelihood that this would happen by chance is quite small in environments with high mixing times in which the transitions are highly biased towards the transient distribution. As such, off-policy correction or model-based approaches are preferable in the general case. That said, in an environment that passive task switches that are independent of the agent's behavior, maintaining a replay buffer with reservoir sampling \citep{MER} will converge in the long-run to the proper steady-state distribution over tasks, only leaving the within task state distribution in need of potential off-policy correction. This insight may shed light on the success of replay based strategies for continual RL, that have been largely tested in settings similar to Example 1 in the main text, particularly considering how they often perform relatively well when it comes to converged final performance.   

\textbf{Definition of Tasks:} In Section 2.2 
we provide a useful definition of tasks for the purposes of our paper. It is largely inline with the literature that often considers tasks as an unobserved component of the state space. For example, in Bayes-Adaptive MDPs unobserved tasks are sampled every episode from a stationary distribution \citep{duff2002optimal,BAMDP,varibad}. On the other hand, Hidden-Mode MDPs formalize MDPs where tasks evolve based on a Markov chain that the agent can only passively observe and not influence \citep{HMMDP,xie2020deep}. Finally, Mixed Observability MDPs (MOMDPs) \citep{MOMDP} allow for the task space to evolve with transition dynamics that involve the agent's behavior. We can even consider decentralized multi-agent settings with learning agents as a special case of the MOMDP setting \citep{lola,CMAML,metamapg}.  
All of these task specifications fit squarely within our framework, but also we do not actually rely on the task being unobserved to highlight the difficulties associated with high mixing times and do not need a formal notion of a task space for the results in our paper to hold. 

\textbf{Tasks That Are Not Explicit:} Indeed, similar concepts to tasks may emerge naturally based on the environment structure alone as is common in Factored MDPs \citep{kearns1999efficient,boutilier2000stochastic, strehl2007efficient,osband2014near,chitnis2020camps,abdulhai2021context}. 
Moreover, agents may learn to decompose the problem into sub-tasks or options \citep{options} on their own. 
While some work does exist that models task evolution as a truly non-stationary process \citep{nonstationaryrlenvs,lecarpentier2019non,chandak}, as pointed out by \citep{crlsurvey}, these models must be very conservative to allow for this non-stationarity and are less likely to be able to exploit regularities between tasks and within task evolution structure as a result.

\begin{figure}[t]
    \centering
    \includegraphics[scale=0.45]{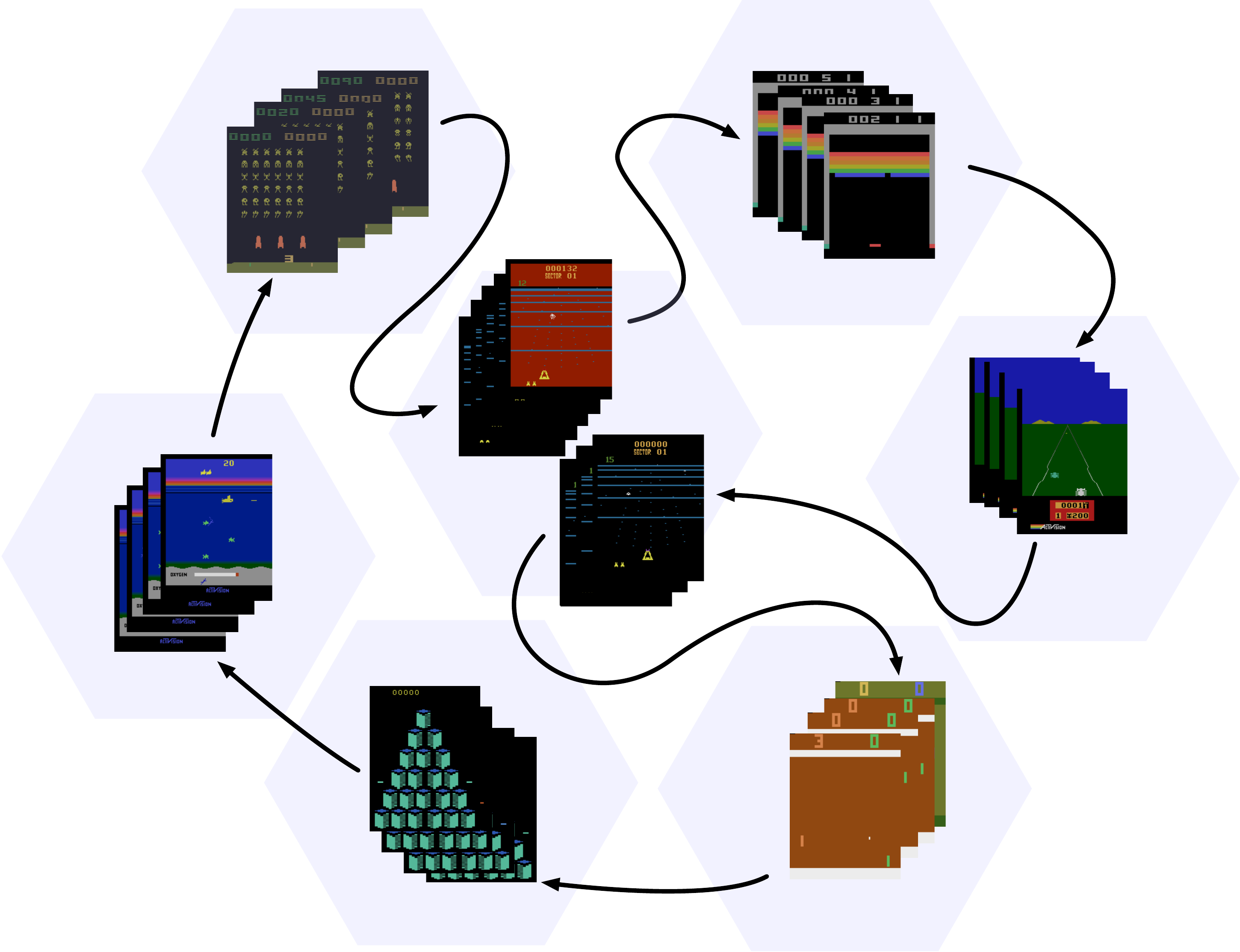}
    \caption{\textbf{Task Structure in Atari Experiments:} We depict an arbitrarily selected random path through the seven Atari games considered in our experiments (each task represented by an hexagonal region in the figure). In this specific instantiation, Beam Rider (center) was played twice before each task was visited at least once. Notice how the maximum path length between two tasks must scale at least with the number of tasks as this number is grown.}
    \label{fig:intuitions}
\end{figure}

\textbf{Illustrating Task Structure in our Experiments}: 
 Figure~\ref{fig:intuitions} of this appendix represents a simple instantiation of the seven tasks (Atari games) of our experiments.
The seven tasks correspond to regions as described in Figure 2 of the main text, and are represented here as frame sequences from the game-play enclosed in hexagonal regions. Each stack of frames has the first frame in the bottom and the last frame up front. The region's boundaries are frames from which the arrows depart; note that these frames connect one region with the \emph{next} (i.e. the arrows always go from the last frame of the current task to the first frame of the next task). Taking them together, the arrows define a specific instantiation of a path through the different regions, which lower bounds the diameter as explained in Figure 2 of the main text.

\subsection{Comparison To Off-Policy Learning}
\textbf{Off-Policy Model-Based:} In particular, our proposed approaches for tackling myopic bias in model-based RL detailed in Appendix \ref{app:expdetails} share similarities with off-policy model-based approaches such as Dyna \citep{sutton1991dyna} and more modern variants for deep RL like MBPO \citep{mbpo}. The main advantage of our approach is that it is able to perform policy improvement directly over the limiting distribution and address the pitfall highlighted by Corollary \ref{corollary:improvement} when performing traditional bootstrapping as these approaches do. Addressing this limitation is important in light of results demonstrating that approaches like Dyna benefit greatly from longer rollouts \citep{holland2018effect}, which highlights the limitations of traditional bootstrapping. Moreover, we consistently achieve lower regret than Dyna across all of our experiments despite allowing for a very large range of rollouts in our hyperparameter search. 
\begin{algorithm}[H]
\caption{$\epsilon$- Return Mixing Time Estimation} 
\label{alg:mixing} 
    \SetKwInput{KwInput}{Input}                
    \SetKwInput{KwOutput}{Output}              
    \DontPrintSemicolon
    \SetKwFunction{Fmix}{MixingTimeEstimation}
    \SetKwProg{Fn}{Function}{:}{}
    \Fn{\Fmix{$env$,$\epsilon$, $\pi$, $|\hat{\cal S}|_{\textrm{max}}$}}{
  
    \tcp{Calculate the asymptotic $h$-step return of the policy $\pi$ }
    \begin{algorithmic}
    \State $s_h \gets env.reset()$  
    \State$G(\pi) \leftarrow 0$
    \State Initialize $\mathcal{H}$ \hfill \tcp{Horizon length for $\rho(\pi)$ calculation} 
    \State \For{h in $1, 2, \dots \mathcal{H}$}{
        \State  $a_h \leftarrow \pi(s_h)$
        \State $s_{h + 1}, r_h \leftarrow env.step(a_h)$
        \State $G(\pi) = G(\pi) + r_h$
        }
    \State \vspace{2mm}$\rho(\pi) = G(\pi) / \mathcal{H}$ \hfill \tcp{Asymptotic h-step return}
    \State \tcp{Store the reward history for each state}
    \State $\textrm{StateRewardHistory} \gets \textrm{dict}()$
    \State $s_h \gets env.reset()$
        \State \For{h in $1, 2, \dots \mathcal{H}$}{
            
            \State   $a_h \leftarrow \pi(s_h)$
            \State $s_{h + 1}, r_h \leftarrow env.step(a_h)$
             \State   $s_{idx} \gets random.randint(1, h)$
             \State \eIf{$s_{idx} < |\hat{\cal S}|_{\textrm{max}}$ }{
                \textrm{StateRewardHistory}[$s_{idx}$] $\gets \left[ \ \right]$
             }{
             \For{$s_i$ in StateRewardHistory}{StateRewardHistory[$s_i$].append($r_h$)}
             }
        }
    \State \tcp{Calculate the mixing time using $\rho(\pi)$ and StateRewardHistory}
    \State $t_{\textrm{ret}}^\pi(\epsilon) \gets \left[ \ \right]$
    \State \For{state in $\textrm{StateRewardHistory}$}{
        \State  rewards $\gets$ StateRewardHistory[state]
        
        \State $t_{\textrm{mix}} \gets \left[ \ \right]$
        \State  $h \gets 0$
        \State $G(h, state, \pi) \gets 0$
        \State \For{r in rewards}{
            \State $h \gets h + 1$
            \State $G(h, state, \pi) \gets G(h, state, \pi) + r$
            \State  $\rho(h, state, \pi) \gets G(h, state, \pi) / h$
            
            \State  \eIf{$|\rho(h, state, \pi) - \rho(\pi)| < \epsilon$}{
            \State  $t_\textrm{mix}.append(h)$
             \;\;}{
            $t_{\textrm{mix}} \gets \left[ \ \right]$
            }
        
        }
        \State $t_{\textrm{ret}}^\pi(\epsilon).append(\min(t_{\textrm{mix}}))$ \hfill \tcp{ Add $\min(t_{\textrm{mix}}) $ to $t_{\textrm{ret}}^\pi(\epsilon)$}
    }
    \textbf{return} Mean($t_{\textrm{ret}}^\pi(\epsilon)$)

\end{algorithmic}
}
\end{algorithm}

\textbf{Off-Policy Model-Free:} A very popular approach to off-policy model-free RL is to leverage \textit{importance sampling} \citep{importancesampling}, which attempts to correct data in an off-policy buffer to better mirror the stationary distribution of the current policy. However, traditional importance sampling faces concerns related to both scalability and variance for the long-horizon and continuing problems that we consider in this work \citep{levine2020offline}. More in the spirit of our work are approaches that consider \textit{marginalized importance sampling}, where agents use dynamic programming in order to estimate the ratio between the stationary and behavior distribution from an off-policy buffer. One group of approaches looking to tackle this problem try to estimate this ratio directly leveraging a Bellman equation style update \citep{hallak2017consistent,gelada2019off,wen2020batch}. These approaches suffer from the same difficulty from Corollary \ref{corollary:improvement} for finding this ratio, making it difficult to successfully apply these approaches in environments with high mixing times. Meanwhile, a second group of approaches attempt to find this ratio by leveraging a learned value function \citep{nachum2019dualdice,nachum2019algaedice,tang2019doubly,nachum2020reinforcement}. These approaches both clearly do not address Corollary \ref{corollary:improvement}, and generally still need expensive optimizations procedures at each step while solving only a regularized (i.e. biased) learning problem. See \citep{levine2020offline} for a comprehensive review of the research highlighted in this paragraph.

\section{Proposed Algorithms} \label{sec:algo-mixing}

\textbf{Estimating The Mixing Time.} In this section, we begin by providing pseudo-code for mixing time estimation as outlined in the main text. For the detailed algorithm please refer to Algorithm \ref{alg:mixing}. Regarding the hyperparameters,  for all our experiments we considered $\cal{H} = \tau \times$ $10^3$. For the experiments reported in Figure 4 and 6, we report the mean and standard deviation across 3 seeds. For the experiments in Figure 5, in order to remove the task bias, we randomly shuffle the list of environments we consider and report the mean and standard deviation across 10 seeds.


    

\textbf{Mixing Time Independent Algorithms for Efficient Scaling.} In the main text we theoretically argued that the polynomial mixing times impact the learning. In this section we will show the empirical evidence on some continual gridworld domains to support those arguments. In order to efficiently learn in the environments with polynomial mixing times, we need to build algorithms which scale in terms of compute independent of the mixing times. While it is challenging to do so in complex domains like Atari, it is an achievable target in tabular domains where we can estimate the steady state distribution of a given policy $\mu^\pi(s)$ in a closed form. More precisely, the steady state equation $\sum_{s \in \cal S} \mu^\pi(s) \sum_{a \in \cal A} \pi(a|s) T(s'|s,a) = \mu^\pi(s') \;\; \forall s' \in \cal S$ can be viewed as a system of equations which can be solved
exactly by performing matrix inversion of a $\cal S \times \cal S$ matrix. Using well known linear algebra libraries, we can then calculate the steady state distribution in closed form and perform sampling from this distribution directly for learning. Here we propose two such algorithms which leverage this system of equations style approach to perform policy improvement with respect to the average reward based on our current model of the environment. We first propose \textbf{on-policy $\rho$ learning} which is similar to on-policy q-learning in that optimal actions are considered as we enter each state based on our current approximation of the average reward per step resulting from
each action. Next we propose \textbf{off-policy $\rho$ learning}, an off-policy version of it where we update the current policy based on the trajectories collected by some behaviour policy. Please refer to Algorithm \ref{alg:on-policy} and Algorithm \ref{alg:off-policy} for more details.

\begin{algorithm}[t]
\caption{On-Policy $\rho$-Learning} 
\label{alg:on-policy} 
    \SetKwFunction{Fon}{OnPolicy$\rho$Learning}
    \SetKwProg{Fn}{Function}{:}{}
    \Fn{\Fon{$env,\epsilon$}}{
    \begin{algorithmic}
  \State \emph{Initialize $\pi$, $\hat{T}$ and $\hat{R}$}
  \State $s_t \gets env.reset()$
  \State \While{\text{not done}}{
     \State  $p \sim uniform([0,1])$
     \State  \eIf{$p \geq \epsilon$:}{
     \State  \For{$a \in \mathcal{A}$:}{
     \State  Solve for $\hat{\mu}^{\pi_m(s_t,a,\pi)}$ using $\sum_{s \in \cal S} \hat{\mu}^\pi(s) \sum_{a \in \cal A} \pi(a|s) T(s'|s,a) = \hat{\mu}^\pi(s')$\;
     \State Solve for $\hat{\rho}(\pi_m(s_t,a,\pi))$ using $\hat{\rho}(\pi) = \sum_{s \in \cal S}  \hat{\mu}^\pi(s) \sum_{a \in \cal A} \pi(a|s) \hat{R}(s,a)$
     \State $a_t = \argmax_{a \in \cal A} \hat{\rho}(\pi_m(a,s_t,\pi))$
     \State Update to greedy policy: \; $\pi = \pi_m(a_t,s_t,\pi)$}}
     {\State Random exploration: \; $a_t \sim uniform(a \in \cal A)$
     \State $s_{t+1}, r_t \gets env.step(s_t, a_t)$
     \State  \emph{Update $\hat{T}$ on $(s_t,a_t,s_{t+1})$ and $\hat{R}$ on $(s_t,a_t,r_t)$}
     \State  Update to next state: \; $s_t = s_{t+1}$
     }
     }
  \State \textbf{return} $\pi,\hat{T},\hat{R}$
  \end{algorithmic}
}
\end{algorithm}

\begin{algorithm}[t]
\caption{Off-Policy $\rho$-Learning} \label{alg:off-policy} 
\SetKwFunction{Foff}{OffPolicy$\rho$Learning}
    \SetKwProg{Fn}{Function}{:}{}
    \Fn{\Foff{$env,\epsilon$, $B$}}{
    \begin{algorithmic}
  \State \emph{Initialize $\pi$, $\hat{T}$ and $\hat{R}$}
  \State $s_t \gets env.reset()$
  \State \While{\text{not done}}{
     \State  $p \sim uniform([0,1])$
     \State \eIf{if $p \geq \epsilon$}{
     \State  Sample an action: \; $a_t \sim \pi(s_t)$}
    {
     \State  Random exploration: \; $a_t \sim uniform(a \in \cal A)$
     }
     \State  $s_{t+1}, r_t \gets env.step(s_t, a_t)$
     \State\emph{Update $\hat{T}$ on $(s_t,a_t,s_{t+1})$ and $\hat{R}$ on $(s_t,a_t,r_t)$ }
     \State  Update to next state: \; $s_t = s_{t+1}$ 
     \State \For{ $i \in [0, ..., B-1]$}{
     \State  $s_i \sim uniform(s \in \cal S)$
     \State \For{ $a \in \mathcal{A}$:}{
     \State  Solve for $\hat{\mu}^{\pi_m(s_t,a,\pi)}$ using $\sum_{s \in \cal S} \hat{\mu}^\pi(s) \sum_{a \in \cal A} \pi(a|s) T(s'|s,a) = \hat{\mu}^\pi(s')$\;
     \State Solve for $\hat{\rho}(\pi_m(s_t,a,\pi))$ using $\hat{\rho}(\pi) = \sum_{s \in \cal S}  \hat{\mu}^\pi(s) \sum_{a \in \cal A} \pi(a|s) \hat{R}(s,a)$
     \State $a_i = \argmax_{a \in \cal A} \hat{\rho}(\pi_m(a,s_i,\pi))$
     \State  Update to greedy policy: \; $\pi = \pi_m(a_i,s_i,\pi)$
     }
     }
     }
  \State \textbf{return} $\pi,\hat{T},\hat{R}$
  \end{algorithmic}
}
\end{algorithm}
\section{Additional Implementation Details and Tabular Experiments} \label{app:expdetails}

In this section we begin by discussing additional details for about our experiments in the main text and then outline our tabular experiments highlighting how approaches that estimate the steady-state distribution directly can combat myopic bias in the presence of polynomial mixing times. 

\subsection{Additional Details for Motivating 3-D Grid Experiments}

Our implementation of REINFORCE was implemented in Pytorch with a learning rate of 0.1, entropy regularization with strength 0.1, and the SGD optimizer. The gradient norm was clipped to a maximum range of 10.0. Additionally, we set the discount factor $\gamma=0$. This implies that each action was associated with only the immediate reward as its expected return. Given the nature of rewards in the problem, this is the optimal choice of $\gamma$ and other values would only frustrate learning further.  

\subsection{Additional Details for Mixing Time Experiments}

\textbf{Compute:} Since our mixing time analysis is based on pretrained policies, our experiments are not GPU heavy. However, to get the accurate mixing time estimates we run the experiments for a large number of asymptotic steps $\mathcal{H} = 1000 \tau$ (at least on the order of a million steps) and the experiments take a long time to run. For small $\tau$ values the experiments took significantly less than 24hrs but for larger values of $\tau$ the experiments took more than 48hrs. Our experiments were deployed on a cluster of Intel x86 machines requesting 1 GPU (either K40, K80, RTX800 or V100s) and 1 CPU for each experiment. RAM was allocated as appropriate for each experiment with larger values of $\mathcal{H}$ requiring more than 100GB. 

For tabular experiments, since all the baselines and our proposed methods are tabular methods, training these models did not need a huge
amount of compute. However, for steady-state approximation using
exact solution methods, we have provided an option of using a pytorch based system of equations solver in the codebase to leverage the fast GPU computations for larger matrices. All the relevant libraries and frameworks used to run our experiments are detailed in the README file provided with our code.

\subsection{Tabular Experiments} 

\begin{table*}[t]
\centering
\resizebox{\textwidth}{!}{
\begin{tabular}{cc||cc||cccc} 
 \toprule
 \textit{Grid Length $d$} & \textit{Steps} & \textit{On-Policy Q-Learning} & \textit{On-Policy $\rho$-Learning} & \textit{Off-Policy Q-Learning} &\textit{Dyna Q-Learning} &\textit{Model-based n-step TD} & \textit{Off-Policy $\rho$-Learning} \\
 \hline
  \multirow{3}{*}{5} & 10k &\g{0.113}{0.028} &\highlight{\g{0.054}{0.003}} &\g{0.100}{0.058} &\g{0.097}{0.046} &\g{0.135}{0.029}&\highlight{\g{0.041}{0.003}} \\
  & 100k &\g{0.081}{0.028} &\highlight{\g{0.048}{0.004}} &\g{0.077}{0.028} &\g{0.101}{0.011}&\g{0.117}{0.024} &\highlight{\g{0.036}{0.003}} \\
 \hline
  \multirow{3}{*}{25} & 10k &\g{0.062}{0.037} &\highlight{\g{0.033}{0.004}} &\g{0.058}{0.039} &\g{0.061}{0.038} &\g{0.060}{0.037}&\highlight{\g{0.057}{0.015}}  \\
 & 100k &\g{0.057}{0.040} &\highlight{\g{0.016}{0.002}} &\g{0.056}{0.042} &\g{0.060}{0.038} &\g{0.059}{0.037}&\highlight{\g{0.019}{0.002}}  \\
  \bottomrule

\end{tabular}
}
\caption{Accumulated lifelong regret per step obtained by an agent in a scalable MDP featuring spatial scaling (Example \ref{def:example1}).}
\label{tab:example1}
\end{table*}
\begin{table*}[t]
\begin{minipage}{\textwidth}
\centering
\resizebox{1.0\columnwidth}{!}{
\scalebox{0.5}
{\begin{tabular}{ccc||cc||cccc} 
 \toprule
 \textit{No. of Rooms $N$} &\textit{Task type} & \textit{Steps} & \textit{On-Policy Q-Learning} & \textit{On-Policy $\rho$-Learning} & \textit{Off-Policy Q-Learning} &\textit{Dyna Q-Learning} & \textit{Model-based  n-step TD} & \textit{Off-Policy $\rho$-Learning}\\
 \hline
  \multirow{6}{*}{4} & \textit{Random}& 10k &\g{0.367}{0.138} &\highlight{\g{0.162}{0.040}} &\g{0.396}{0.230} &\g{0.300}{0.124} &\g{0.274}{0.032}&\highlight{\g{0.180}{0.038}} \\
 
  
  & & 100k &\g{0.279}{0.059} &\highlight{\g{0.152}{0.038}} &\g{0.242}{0.090} &\g{0.222}{0.049} &\g{0.201}{0.032}&\highlight{\g{0.153}{0.040}}  \\
   
   

   
  & \textit{Cycles}& 10k &\g{0.271}{0.115} &\highlight{\g{0.074}{0.030}} &\g{0.287}{0.161} &\g{0.171}{0.070}  &\g{0.229}{0.075}&\highlight{\g{0.099}{0.033}}  \\
 
  & & 100k &\g{0.165}{0.056} &\highlight{\g{0.063}{0.029}} &\g{0.120}{0.045} &\g{0.120}{0.045} &\g{0.130}{0.042} &\highlight{\g{0.065}{0.029}}  \\

 \hline
 \multirow{6}{*}{16} & \textit{Random}& 10k &\g{0.365}{0.230} &\highlight{\g{0.138}{0.014}} &\g{0.348}{0.178} &\g{0.410}{0.087} &\g{0.355}{0.105}&\highlight{\g{0.292}{0.030}}  \\
 
  
  & & 100k &\g{0.303}{0.092} &\highlight{\g{0.091}{0.009}} &\g{0.321}{0.095} &\g{0.187}{0.017} &\g{0.334}{0.030}&\highlight{\g{0.106}{0.010}} \\
  
  
  
  
  & \textit{Cycles}& 10k &\g{0.338}{0.162} &\highlight{\g{0.100}{0.026}} &\g{0.303}{0.152} &\g{0.364}{0.084} &\g{0.412}{0.114}&\highlight{\g{0.287}{0.041}} \\
 
 
  & & 100k &\g{0.243}{0.067} &\highlight{\g{0.062}{0.017}} &\g{0.363}{0.122} &\g{0.144}{0.028} &\g{0.181}{0.04}&\highlight{\g{0.083}{0.017}}  \\
  
   
  
 
  
 
  \hline
   \multirow{3}{*}{2} & \textit{Curricular}& 10k &\g{0.452}{0.128} &\highlight{\g{0.343}{0.071}} &\g{0.379}{0.0.095} &\g{0.400}{0.086}&\g{0.408}{0.096} &\highlight{\g{0.354}{0.080}} \\
 
  & & 100k &\g{0.424}{0.109} &\highlight{\g{0.340}{0.067}} &\g{0.362}{0.098} &\g{0.366}{0.092} &\g{0.384}{0.087}&\highlight{\g{0.340}{0.070}} \\
 \hline
 \multirow{3}{*}{3} & \textit{Curricular}& 10k  &\g{0.359}{0.157} &\highlight{\g{0.281}{0.043}} &\g{0.389}{0.234} &\g{0.322}{0.091}&\g{0.331}{0.118} &\highlight{\g{0.260}{0.055}}  \\
 
  & & 100k  &\g{0.306}{0.097} &\highlight{\g{0.259}{0.073}} &\g{0.306}{0.097}&\g{0.300}{0.076} &\g{0.285}{0.076} &\highlight{\g{0.250}{0.063}}\\
  \hline
 \multirow{3}{*}{4} & \textit{Curricular}& 10k  &\g{0.283}{0.059} &\highlight{\g{0.180}{0.056}} &\g{0.311}{0.094} &\g{0.246}{0.042} &\g{0.286}{0.074}&\highlight{\g{0.195}{0.071}} \\
 
  & & 100k &\g{0.245}{0.069} &\highlight{\g{0.161}{0.053}} &\g{0.207}{0.037} &\g{0.210}{0.045} &\g{0.209}{0.055}&\highlight{\g{0.183}{0.076}} \\
 
  \bottomrule
\end{tabular}
}
}
\caption{Accumulated lifelong regret per step obtained by an agent in a scalable MDP featuring bottleneck scaling (Example \ref{def:example2}). The values shown are for the three room transition variants across different $N$ values with each room of size $d=5$. }
\label{tab:example2-random-cycles}
\end{minipage}
\end{table*}

\begin{table*}[t]
\begin{minipage}{\textwidth}
\centering
\renewcommand{\arraystretch}{1.25}
\resizebox{\textwidth}{!}{
\scalebox{0.5}
{\begin{tabular}{cc||cc||cccc} 
 \toprule
 \textit{Exponent $x$} & \textit{Steps} & \textit{On-Policy Q-Learning} & \textit{On-Policy $\rho$-Learning} & \textit{Off-Policy Q-Learning} &\textit{Dyna Q-Learning} &\textit{Model-based n-step TD} & \textit{Off-Policy $\rho$-Learning}\\
 \hline
  \multirow{3}{*}{2} & 10k &\g{0.245}{0.016} &\highlight{\g{0.125}{0.008}} &\g{0.286}{0.024} &\g{0.240}{0.018} &\g{0.279}{0.025}&\highlight{\g{0.216}{0.010}}  \\
 
   & 100k &\g{0.243}{0.013} &\highlight{\g{0.062}{0.002}} &\g{0.233}{0.020} &\g{0.124}{0.012}&\g{0.201}{0.012} &\highlight{\g{0.070}{0.002}} \\
 \hline
 \multirow{3}{*}{3} & 10k &\g{0.233}{0.024} &\highlight{\g{0.089}{0.005}} &\g{0.253}{0.024} &\g{0.262}{0.012} &\g{0.262}{0.012}&\highlight{\g{0.252}{0.016}}  \\
 
   & 100k &\g{0.197}{0.019} &\highlight{\g{0.049}{0.003}} &\g{0.194}{0.022} &\g{0.121}{0.011} &\g{0.168}{0.013}&\highlight{\g{0.072}{0.003}}  \\
  \hline
 \multirow{3}{*}{4} & 10k &\g{0.220}{0.036} &\highlight{\g{0.066}{0.007}} &\g{0.243}{0.046} &\g{0.269}{0.085} &\g{0.243}{0.051}&\highlight{\g{0.236}{0.034}}  \\
 
   & 100k &\g{0.185}{0.014} &\highlight{\g{0.047}{0.003}} &\g{0.168}{0.014} &\g{0.140}{0.012} &\g{0.163}{0.015}&\highlight{\g{0.089}{0.007}}  \\
 
  \bottomrule
\end{tabular}
}
}
\caption{Accumulated lifelong regret per step obtained by an agent in a scalable MDP featuring cycle length scaling (Example \ref{def:example3}). The values shown are for the \textit{cyclic} room transitions with $N=16$ rooms. }
\label{tab:example3}
\end{minipage}
\end{table*}

To augment the scalable MDP use cases for deep continual RL highlighted in example 1 in the main text, we consider three grid world based examples with regions of size $d\times d$ in this section to outline some of the key ways that scaling often contributes to polynomial mixing times in practice. 
We then empirically analyze the scaling behavior of Algorithms \ref{alg:on-policy} and \ref{alg:off-policy} on these environment classes using accumulated lifelong regret per step and contrast this performance against relevant baselines. We perform a grid search over learning rate, exploration parameter $\epsilon$ and batchsize $B$ and pick the best hyperparameters for all models (including the baselines). For each experiment we report the mean and the standard deviation across 10 seeds. 

\begin{example}
(Spatial dimensions, $d$): In this episodic task, the agent is placed at an arbitrary location in a $d\times d$ grid world and must reach a goal in an arbitrary location that is fixed across episodes. The agent is only rewarded upon reaching the goal location, which implies that the expected diameter of $\pi^*$ is $\mathbb{E}[D^{\pi^*}] \in \Omega(d)$. Since $|\mathcal{S}|=d^2$ for this class, we have $\mathbb{E}[D^{\pi^*}] \in \Omega(|\mathcal{S}|^{\frac{1}{2}})$. 
\label{def:example1}
\end{example}

Results for $d=5,25$ are shown in Table \ref{tab:example1}. Our proposed algorithms consistently outperform baseline models in terms of lifelong regret. 


\begin{example}
\label{def:example2}
(Number of Bottlenecks, $N$): Consider $N$ grid world regions $\{\mathcal{R}_i\}_{i=1}^N$, implying $|\mathcal{S}|=Nd^2$. Each has an arbitrary starting location when entering from the previous region and an arbitrary goal location serving as a bottleneck transporting the agent to the next region. The agent is only rewarded when it reaches the goal location of its current region, which implies that $\mathbb{E}[D_{\mathcal{R}_i}^{\pi^*}] \in \Omega(d)$ for all $i$. We consider three possibilities for how regions are connected. 
    \textit{Cycle Transitions:} the regions are accessed in a strict order with no repeats. We know that $\mathbb{E}[D^{\pi^*}] \in \Omega(d \times N)$, so if $N$ is scaled with $d$ fixed, $\mathbb{E}[D^{\pi^*}] \in \Omega(|\mathcal{S}|)$. 
    \textit{Random Transitions:} the regions are accessed randomly with repeats. We again know that $\mathbb{E}[D^{\pi^*}] \in \Omega(d \times N)$, so again $\mathbb{E}[D^{\pi^*}] \in \Omega(|\mathcal{S}|)$.
    \textit{Curricular Transitions:} the regions are accessed in a curricular fashion i.e. $\mathcal{R}_1,\mathcal{R}_1,\mathcal{R}_2,\mathcal{R}_1,\mathcal{R}_2,\mathcal{R}_3,$ \textit{etc.} We know $\mathbb{E}[D^{\pi^*}] \in \Omega(d \times N!)$, so if $N$ is scaled and $d$ is fixed, $\mathbb{E}[D^{\pi^*}] \in \Omega(|\mathcal{S}|^{N-1})$. 
\end{example}

Results for $N = 4, 16$ and $d = 5$ are shown in Table \ref{tab:example2-random-cycles}. Our models consistently outperform baselines for both cyclic and random transitions regardless of the number of rooms. 
The curricular transition case is challenging since the diameter scales exponentially with $N$, making the diameter substantially higher. The corresponding convergence rates of all methods (Table \ref{tab:example2-random-cycles}) are lower as expected by regret bounds \citep{jaksch2010near}. Nevertheless, our proposed models still outperform the baselines.

\begin{example}
\label{def:example3}
(Cycle length, $\tau$): $N$ grid world regions with cyclic transitions for which the agent is expected to transition between regions after every $\tau$ environment steps. We use cyclic transitions to highlight similarities to typical settings in continual RL \citep{crlsurvey}. Here, room transitions are passive: the agents' current policy has no direct effect on the room transitions which purely depend on $\tau$.  $\tau \geq 2d$ as otherwise some regions would be impossible to solve, so we can assume a form $\tau = cd^x \;\; \forall c \geq 2, x \geq 1$. This implies that as we scale $x$ keeping $c, d,$ and $N$ constant, $\mathbb{E}[D^{\pi^*}] \in \Omega(|\mathcal{S}|^{x/2})$.
\end{example}

Note that this example bears strong similarity to Example 1 discussed in the main text and can be seen as the grid world analog of that setting. Results for $x = 2, 3$ and $4$ keeping $d = 5$ and $N = 16$ fixed are shown in  Table \ref{tab:example3}. As expected, our proposed methods achieve better sample efficiency.





\section{Detailed Proofs For Key Results} \label{app:proofs}

In this section, we provide detailed proofs for all of the propositions and theorems in our paper following the order that they are presented in the main text. For each proof we first remind readers of the main result and provide a proof sketch before detailing how each step is achieved. 

\setcounter{subsection}{-1}
\subsection{Formal Description of MDP Scaling} 

In this section, we formalize a scalable family of MDPs $\mathbb{C}_\sigma$, using a scaling function, $\sigma$. 
We consider MDPs with state space $\mathcal{S}$ parametrized by an $n$-dimensional vector $\bm{q}\in\mathbb{R}^n$. 
In general, $\bm{q}$ can contain spatial components determining spatial properties of $\mathcal{S}$ (e.g. spatial dimensions of a grid world), as well as non-spatial components that determine qualitative features (e.g. number of `cherry' states in a rewarded sequence of target states in a grid world). As a continuous parametrization of the MDP, $\bm{q}$ serves as a useful target for formalizing scaling. %
We formulate scaling using the following definition:

\textbf{Definition:} A \textit{scaling function}
 is a continuous deformation $\sigma:\mathbb{R}^n\times\mathbb{R}\to\mathbb{R}^n$ parametrized by a scalar $\nu \in \mathbb{R}$ that takes any $\bm{q}_0$ to $\bm{q}_\nu=\sigma(\bm{q}_0,\nu)$, with $\sigma(\cdot,0)$ as the identity map. 
 
We denote by $\Sigma$ the set of all scaling functions on $\bm{q}$.
Thus, each $\sigma\in\Sigma$ induces $\mathbb{C}_\sigma=\{\mathcal{M}_\nu\}$, a $\nu$-parametrized abstract class of MDPs along a smooth path in $q$-space.
We consider scaling functions that scale up the MDP:

\textbf{Definition:} An
\textit{expansive scaling function} is a scaling function for which
$\partial q_{\nu,i}/\partial\nu\geq0$ for all $i=1, \dots, n$, and $\partial q_{\nu,i}/\partial\nu>0$ for at least one $i$.

Thus, the MDP will never decrease in size along a path in $\bm{q}$-space taken through an expansive scaling function. An important example of expansive scaling functions is:

\textbf{Definition:} A \textit{proportional scaling function} is an expansive scaling function
taking the form $\sigma(\bm{q},\nu)=\bm{q}+\nu\Delta\bm{q}$, where $\Delta\bm{q}\in\mathbb{R}^n$ and $\Delta q_i$ is the linear rate of grow of $q_{\nu,i}$ with $\nu$.

We focus on proportional scaling functions in our definition of scalable MDPs stated again in \ref{sec:thm1}.




\subsection{Proof of Proposition 1} 

\textbf{Proposition 1} \textit{If all MDPs $\mathcal{M}$ within the subclass of MDPs $\mathbb{C}$ have $t_\textnormal{ret}^{\pi^*}$, $t_\textnormal{ces}^{\pi^*}$, $t_\textnormal{mix}^{\pi^*}$, $D^{\pi^*}$, or $D^* \in \Omega(|\mathcal{S}|^k)$ for some $k>0$ we can say that $\mathbb{C}$ has a polynomial mixing time.} 

Proposition 1 is thus reliant on the following definition of a \textit{polynomial mixing time}: 

\textbf{Definition 2:}\textit{
A set or family of MDPs $\mathbb{C}$ %
has a \textbf{polynomial mixing time} if the environment mixing dynamics contributes a $\Omega(|\mathcal{S}|^k)$ multiplicative increase for some $k>0$ to the intrinsic lower bound on regret %
as $|\mathcal{S}| \rightarrow \infty \;\; \forall \mathcal{M} \in \mathbb{C}$.}



\textbf{Proof Sketch:} It holds directly from \cite{jaksch2010near} Theorem 5 that following Assumption 1 
from the main text yields a lower bound on regret for RL algorithms $\textrm{Regret}(H) \in \Omega(\sqrt{D^*|\mathcal{S}||\mathcal{A}|H})$, which implies that if $D^* \in \Omega(|\mathcal{S}|^k)$ for some $k>0$, there must be a \textit{polynomial mixing time}. We then begin by noting how this regret bound holds equally for $D^{\pi^*}$. We proceed to demonstrate that if $t_\textnormal{mix}^{\pi^*}$ is polynomial in $|\mathcal{S}|$, the same must be true for $D^{\pi^*}$. We then establish the relationship between $t_\textnormal{ret}^{\pi^*}$ and $t_\textnormal{mix}^{\pi^*}$ and go on to establish the relationship between $t_\textnormal{ces}^{\pi^*}$ and $t_\textnormal{mix}^{\pi^*}$. As such, it is demonstrated that if any of these metrics have a polynomial dependence, the regret bound must as well. 

\textbf{Analysis for $D^{\pi^*}$:} As discussed extensively in \citep{osband2016lower}, the common approach to proving regret bounds in RL and bandit settings it to develop a counter-example for which you can demonstrate that it is impossible for an algorithm to get below a certain regret as a function of the problem parameters. This is the approach taken by \cite{jaksch2010near} to establish the $\textrm{Regret}(H) \in \Omega(\sqrt{D^*|\mathcal{S}||\mathcal{A}|H})$ bound. However, a careful analysis of their proof reveals that for the problem they consider $D^*=D^{\pi^*}$. As such the proof of Theorem 5 in \citep{jaksch2010near} can equally be used to establish that $\textrm{Regret}(H) \in \Omega(\sqrt{D^{\pi^*}|\mathcal{S}||\mathcal{A}|H})$. Note that since by definition  $\Omega(\sqrt{D^{\pi^*}|\mathcal{S}||\mathcal{A}|H}) \subseteq \Omega(\sqrt{D^*|\mathcal{S}||\mathcal{A}|H})$, the more general result is for $D^*$. Moreover, if this were not the case it would invalidate the lower bound presented by \citep{kearns2002near} in terms of $t_\textnormal{ret}^{\pi^*}$. As such, it is clear that we cannot simply ignore the size of the problem from the perspective of the optimal policy $\pi^*$ when conducting regret analysis. 

\textbf{Analysis for $t_\textnormal{mix}^{\pi^*}$:} It is well known that in bounded degree transitive graphs the mixing time $t_\textnormal{mix} \in O(D^3)$ where $D$ is the graph diameter \citep{mixingtimesbook}. Note, however, that it is widely conjectured to be at most $t_\textnormal{mix} \in O(D^2)$ \citep{mixingtimesbook}. Taking the more general case, this implies then that for any policy $\pi$ and MDP $\mathcal{M}$, $D^\pi \in \Omega((t^\pi_\textnormal{mix})^{1/3})$. Therefore, if $t^{\pi^*}_\textnormal{mix} \in \Omega(|\mathcal{S}|^k)$ for some $k > 0$, then the regret must be in $\textrm{Regret}(H) \in \Omega(\sqrt{|\mathcal{S}|^{k/3}|\mathcal{S}||\mathcal{A}|H})$. Therefore, Definition 2 
is satisfied. 

\textbf{Analysis for $t_\textnormal{ret}^{\pi^*}$:} It was established in Lemma 1 of \citep{kearns2002near} that $t_\textnormal{mix}^{\pi} \in \Omega(t_\textnormal{ret}^{\pi})$ for any policy $\pi$. So this then implies that $t_\textnormal{mix}^{\pi^*} \in \Omega(t_\textnormal{ret}^{\pi^*})$. Therefore, we can follow the result from the last paragraph to see that if $t^{\pi^*}_\textnormal{ret} \in \Omega(|\mathcal{S}|^k)$ for some $k > 0$, it must also be true that Definition 2 is satisfied. 

\textbf{Analysis for $t_\textnormal{ces}^{\pi^*}$:} It has been established that $7t_\textnormal{mix} \geq t_\textnormal{ces}$ for any finite chain \citep{mixingtimesbook}. Therefor, $t^{\pi^*}_\textnormal{mix} \in \Omega(t^{\pi^*}_\textnormal{ces})$. So by extension of our results presented above, if  $t^{\pi^*}_\textnormal{ces} \in \Omega(|\mathcal{S}|^k)$ for some $k > 0$, Definition 2 must hold.





\subsection{Proof of Proposition 2}

\textbf{Proposition 2:}
\textit{Any scalable MDPs $\mathbb{C}_\sigma$ exhibits a polynomial mixing time if there exists a scalable region $\mathcal{R}_\nu$ 
such that $\mathbb{E}_{\mu^{\pi^*}}[t^{\pi^*}_{\mathcal{R}_\nu}] \in \Omega(|\mathcal{S}|^{k})$ for some  $k > 0$.} 

\textbf{Proof Sketch:} We first present the bottleneck ratio of a region. We then demonstrate the relationship connecting the bottleneck ratio to the residence time. We conclude by establishing the connection between the bottleneck ratio and the mixing time of the optimal policy. This then allows us to connect the residence time of the optimal policy to its mixing time, which allows us to establish a polynomial mixing time leveraging the result from Proposition 1. 

We begin by defining an edge measure (or ergodic flow) as $\xi^\pi(s'|s):=\mu^\pi(s)T^\pi(s'|s)$ \citep{mixingtimesbook}. For reasoning about regions of state space $\mathcal{R}$, we can further define an edge measure as $\xi^\pi(\mathcal{S}\setminus\mathcal{R}|\mathcal{R}) = \sum_{s' \in \mathcal{S}\setminus\mathcal{R}} \sum_{s \in \mathcal{R}} \xi^\pi(s'|s)$ and steady-state probability $\mu^\pi(\mathcal{R})=\sum_{s \in \mathcal{R}} \mu^\pi(s)$. This can then be used to define the \text{bottleneck} ratio of the set $\mathcal{R}$ \citep{mixingtimesbook}:

\begin{equation} \label{eq:bottleneckratio}
\mho^\pi(\mathcal{R}):=\frac{\xi^\pi(\mathcal{S}\setminus\mathcal{R}|\mathcal{R})}{\mu^\pi(\mathcal{R})} = \frac{1}{\mathbb{E}_{\mu^{\pi}}[t^\pi_\mathcal{R}]}
\end{equation}
With the last equality we note the relationship between the bottleneck ratio of a region $\mathcal{R}$ and the \textit{expected residence time} $\mathbb{E}_{\mu^{\pi}}[t^\pi_\mathcal{R}]$, which is the expected probability of being in $\mathcal{R}$ divided by the expected probability of exiting $\mathcal{R}$. We can finally now define the bottleneck ratio of the entire Markov chain $T^\pi(s'|s)$ (also called the conductance or Cheeger constant) as $\mho_*^\pi:= \min_{\mathcal{R} \subset \mathcal{S} : \mu^\pi(\mathcal{R}) \leq 1/2 } \mho^\pi(\mathcal{R})$ \citep{mixingtimesbook}. Importantly the bottleneck ratio can be used to both upper bound and lower bound on the mixing time \citep{mixingtimesbook}. For our purposes, we will primarily utilize the fact that $t_\textrm{mix}^{\pi}(1/4) \geq 1/4\mho_*^\pi$ as stated in Theorem 7.4 of \citep{mixingtimesbook}. We can then consider the context of $\pi^*$:

\begin{equation} \label{eq:mixingtimebound}
\begin{split}
t_\textrm{mix}^{\pi^*}(1/4) &\geq \frac{1}{4\mho_*^{\pi^*}} =  \max_{\mathcal{R} \subset \mathcal{S} : \mu^{\pi^*}(\mathcal{R}) \leq 1/2} \frac{1}{4\mho^{\pi^*}(\mathcal{R})} \\
&= \max_{\mathcal{R} \subset \mathcal{S} : \mu^{\pi^*}(\mathcal{R}) \leq 1/2} \frac{\mathbb{E}_{\mu^{\pi^*}}[t^{\pi^*}_\mathcal{R}]}{4}
\end{split}
\end{equation}

This in turn implies the following result for the conventional mixing time $t_\textrm{mix}^{\pi^*}(1/4)$:

\begin{equation}
t_\textrm{mix}^{\pi^*}(1/4) \in \Omega(\mathbb{E}_{\mu^{\pi^*}}[t^{\pi^*}_\mathcal{R}]) \;\; \forall \mathcal{R} \subset \mathcal{S} : \mu^{\pi^*}(\mathcal{R}) \leq 1/2
\end{equation}

This result is nearly what we are looking for. However, we would like to express the mixing time at an arbitrary precision $\epsilon$ in terms of the residence time rather than only considering $\epsilon=1/4$. To achieve this we note that the proof of Theorem 7.4 in \citep{mixingtimesbook} can easily be extended for arbitrary $\epsilon$ with $\epsilon=1/4$ only being used as a convention in the literature. Taking the results presented after equation 7.10 in \citep{mixingtimesbook} and rearranging them with our notation, we find that:

\begin{equation}
    t_\textrm{mix}^{\pi^*}(\epsilon) \geq \frac{1-\epsilon-\mu^{\pi^*}(\mathcal{R})}{\mho_*^{\pi^*}}
\end{equation}

As a result as long as we restrict $\mu^{\pi^*}(\mathcal{R}) \leq 1 - \epsilon - \delta$ for some $0 < \delta < 1-\epsilon$ we can conclude that $t_\textrm{mix}^{\pi^*}(\epsilon) \geq \delta t^{\pi^*}_\mathcal{R}$. This implies that we can state the following asymptotic result as long as the inequality on the density is strict, leading to some effective positive $\delta$:

\begin{equation} \label{eq:mainresultprop1}
t_\textrm{mix}^{\pi^*}(\epsilon) \in \Omega(\mathbb{E}_{\mu^{\pi^*}}[t^{\pi^*}_\mathcal{R}]) \;\; \forall \mathcal{R} \subset \mathcal{S} : \mu^{\pi^*}(\mathcal{R}) < 1 - \epsilon
\end{equation}

Therefore, if any scalable region $\mathcal{R}_\nu$ exists where $\mu^{\pi^*}(\mathcal{R}_\nu) < 1 - \epsilon$ and $\mathbb{E}_{\mu^{\pi^*}}[t^{\pi^*}_{\mathcal{R}_\nu}] \in \Omega(|\mathcal{S}|^{k})$ for some  $k > 0$, we also know that $t_\textrm{mix}^{\pi^*}(\epsilon) \in \Omega(|\mathcal{S}|^{k})$ and thus a polynomial mixing time is ensured following Proposition 1. Moreover, we can finalize our proof of Proposition 2 by noting that any arbitrarily large scalable region of density greater than the bound i.e. $\mu^{\pi^*}(\mathcal{R}_\nu) \geq 1 - \epsilon$ that scales polynomially must also contain a scalable sub-region $\mathcal{R'}_\nu \subset \mathcal{R}_\nu$ of arbitrarily small density $\mu^{\pi^*}(\mathcal{R'}_\nu) < 1 - \epsilon$ that scales at least at the same rate. As such, there is no need to provide an upper bound for the steady-state region probability density in Proposition 2.

\subsection{Proof of Theorem 1}\label{sec:thm1}
\textbf{Theorem 1:}\textit{
(Mixing Time Scaling): \;
Any scalable MDP $\mathbb{C}_\sigma$ has a polynomial mixing time.}


Theorem 1 uses the following definitions of \textit{scalable MDP} and \textit{polynomial mixing time}. 

\textbf{Definition 1:}\textit{ A \textbf{scalable MDP} is a family of MDPs $\mathbb{C}_\sigma=\{\mathcal{M}_\nu\}$ 
arising from a proportional scaling function $\sigma$ 
satisfying the property that 
there exists an initial scalable region $\mathcal{R}_0$ with finite interior, $\mu^{\pi^*}(\partial\mathcal{R}_0) < \mu^{\pi^*}(\mathcal{R}_0)$, that scales so that $\mu^{\pi^*}(\partial\mathcal{R}_\nu) < \mu^{\pi^*}(\mathcal{R}_\nu)$ %
as $\nu\to\infty$ and thus $|\mathcal{S}|\to\infty$.}

\textbf{Definition 2:}\textit{ A set or family of MDPs $\mathbb{C}$ %
has a \textbf{polynomial mixing time} if the environment mixing dynamics contributes a $\Omega(|\mathcal{S}|^k)$ multiplicative increase for some $k>0$ to the intrinsic lower bound on regret %
as $|\mathcal{S}| \rightarrow \infty \;\; \forall \mathcal{M} \in \mathbb{C}$.}

\textbf{Proof Sketch:} We begin by further analyzing the bottleneck ratio and residence time introduced in Proposition 2 by connecting it to the relative densities of a region and its boundary. We then consider 
how, under the conditions we consider, a proportionally scaled region must grow faster than its boundary and the region to boundary ratio is lower bounded by the respective increase in their sizes as $\nu$ increases. Finally, we establish that the scaling of some regions of any scalable MDP undergoing proportional scaling is polynomially larger than the scaling of its boundary and prove polynomial mixing through the use of Proposition 1. 

We begin by further analyzing the ergodic flow of any policy $\pi$ in region $\mathcal{R}$ and boundary $\partial \mathcal{R}$: 

\begin{equation}
\begin{split}
 \xi^\pi(\mathcal{S}\setminus\mathcal{R}|\mathcal{R}) = \sum_{s' \in \mathcal{S}\setminus\mathcal{R}} \sum_{s \in \mathcal{R}} \mu^\pi(s)T^\pi(s'|s) \\
 = \sum_{s' \in \mathcal{S}\setminus\mathcal{R}} \sum_{s \in \mathcal{\partial R}} \mu^\pi(s)T^\pi(s'|s) \leq \mu^{\pi}( \partial \mathcal{R})
\end{split}
\end{equation}

As such, we can use this result to provide a lower bound for the expected residence time:

\begin{equation}
    \mathbb{E}_{\mu^{\pi}}[t^\pi_\mathcal{R}] = \frac{\mu^\pi(\mathcal{R})} {\xi^\pi(\mathcal{S}\setminus\mathcal{R}|\mathcal{R})} \geq  \frac{\mu^\pi(\mathcal{R})} {\mu^\pi(\mathcal{\partial R})}
\end{equation}





This in turn implies the following relation to the mixing time by subbing in equation \ref{eq:mainresultprop1}:

\begin{equation}
t_\textrm{mix}^{\pi^*}(\epsilon) \in \Omega(\mathbb{E}_{\mu^{\pi^*}}[t^{\pi^*}_\mathcal{R}]) \in \Omega\bigg(\frac{\mu^{\pi^*}(\mathcal{R})}{\mu^{\pi^*}( \partial \mathcal{R})}\bigg) \;\; \forall \mathcal{R} \subset \mathcal{S} : \mu^{\pi^*}(\mathcal{R}) < 1 - \epsilon
\label{eq:tmixratio}
\end{equation}

From this analysis it is clear that we can characterize polynomial mixing times by understanding how the probability mass on any scalable region and the mass on its boundary scale with the state space. Formally, for some scaling function $\sigma$ with scaling parameter $\nu$ giving a state space $\mathcal{S}_\nu$, we would like to understand how the mass on a scalable region $\mathcal{R}_\nu$ and its boundary, $\partial\mathcal{R}_\nu$, $\mu^{\pi^*}(\mathcal{R}_\nu)$ and $\mu^{\pi^*}(\partial\mathcal{R}_\nu)$, respectively, scale with $|\mathcal{S}_\nu|$. Without loss of generality, we set a reference MDP at $\nu=0$ and denote some region in it as $\mathcal{R}_0$. We suppress the dependence of $\pi^*$ and $\mu^{\pi^*}$ on $\nu$ in our notation for clarity.   
We now derive the main result.
By the definition of a scalable MDP, the bulk-to-boundary mass ratio of a region can not decrease as $\nu$ scales up the state space. Thus,  
\begin{equation}
\frac{\mu^{\pi^*}(\mathcal{R}_\nu)}{\mu^{\pi^*}(\partial\mathcal{R}_\nu)} \ge \frac{\mu^{\pi^*}(\mathcal{R}_0)}{ \mu^{\pi^*}(\mathcal{\partial{R}}_0)}\;.
\end{equation}
Since mass on a scaled region and its boundary cannot increase with $\nu$, $\mu^{\pi^*}(\mathcal{R}_\nu) \leq \mu^{\pi^*}(\mathcal{R}_0)$ and $\mu^{\pi^*}(\partial\mathcal{R}_\nu) \leq \mu^{\pi^*}(\partial\mathcal{R}_0)$. Thus,
\begin{align}
\label{th1ineq}
\frac{\mu^{\pi^*}(\mathcal{R}_\nu)}{\mu^{\pi^*}(\partial\mathcal{R}_\nu)} &\ge \frac{\mu^{\pi^*}(\mathcal{R}_0)}{ \mu^{\pi^*}(\mathcal{\partial{R}}_0)}\cdot \frac{1-\frac{\mu^{\pi^*}(\mathcal{R}_\nu)}{\mu^{\pi^*}(\mathcal{R}_0)}}{1-\frac{\mu^{\pi^*}(\partial\mathcal{R}_\nu)}{\mu^{\pi^*}(\partial\mathcal{R}_0)}}\nonumber \\
\frac{\mu^{\pi^*}(\mathcal{R}_\nu)}{\mu^{\pi^*}(\partial\mathcal{R}_\nu)} &\ge\frac{\mu^{\pi^*}(\mathcal{R}_\nu) - \mu^{\pi^*}(\mathcal{R}_0)}{\mu^{\pi^*}(\partial\mathcal{R}_\nu) - \mu^{\pi^*}(\mathcal{\partial{R}}_0) }\;.
\end{align}

To demonstrate the scaling of the right hand side, we can approximate the numerator and denominator.
For sufficiently large $\mathcal{R}$, the variation of the size of and mass on the region and its boundary vary in an approximately continuous way  with $\nu$ so we can employ the inverse function theorem to expand the mass on a region in its size, $|\mathcal{R}_\nu|$, and the same with its boundary, $|\partial\mathcal{R}_\nu|$.  To first order around $\nu=0$, we have
\begin{align}
\mu^{\pi^*}(\mathcal{R}_\nu) &\approx \mu^{\pi^*}(\mathcal{R}_0) + \frac{\partial \mu^{\pi^*}}{\partial|\mathcal{R}_\nu|}\bigg|_{\nu=0}\Delta{|\mathcal{R}|}\;,\\
\mu^{\pi^*}(\mathcal{\partial{R}}_\nu) &\approx \mu^{\pi^*}(\mathcal{\partial{R}}_0) + \frac{\partial\mu^{\pi^*}}{\partial|\partial\mathcal{R}_\nu|}\bigg|_{\nu=0}\Delta{|\mathcal{\partial{R}}|} \;,
\end{align}
for deviations, $\Delta|\mathcal{R}|:=|\mathcal{R}_\nu|-|\mathcal{R}_0|$ and $\Delta{|\mathcal{\partial{R}}|}:=|\mathcal{\partial{R}}_\nu|-|\mathcal{\partial{R}}_0|$ (note that the notation $\partial/\partial|\partial\mathcal{R}|$ is the partial derivative with respect to the size of the region's boundary, $\partial\mathcal{R}$). 
Then, arranging the terms:
\begin{equation}
\frac{\mu^{\pi^*}(\mathcal{R}_\nu) - \mu^{\pi^*}(\mathcal{R}_0)}{\mu^{\pi^*}(\partial\mathcal{R}_\nu) - \mu^{\pi^*}(\mathcal{\partial{R}}_0) }
\approx \frac{\frac{\partial \mu^{\pi^*}}{\partial|\mathcal{R}_\nu|}\big|_{\nu=0}\Delta{|\mathcal{R}}|}
{\frac{\partial\mu^{\pi^*}}{\partial|\partial\mathcal{R}_\nu|}\big|_{\nu=0}\Delta{|\mathcal{\partial{R}}}|}\;.
\end{equation}
Subbing into equation~\ref{th1ineq} above we get:
\begin{equation}
\frac{\mu^{\pi^*}(\mathcal{R}_\nu)}{\mu^{\pi^*}(\partial\mathcal{R}_\nu)}
\ge \frac{\frac{\partial \mu^{\pi^*}}{\partial|\mathcal{R}_\nu|}\big|_{\nu=0}\Delta{|\mathcal{R}}|}
{\frac{\partial\mu^{\pi^*}}{\partial|\partial\mathcal{R}_\nu|}\big|_{\nu=0}\Delta{|\mathcal{\partial{R}}}|}\;.
\end{equation}

Noting that the derivatives are of equal sign (less than or equal to zero) following Definition \ref{def:scalableMDP}
, we can now state the following scaling dependence:
\begin{equation}
\frac{\mu^{\pi^*}(\mathcal{R}_\nu)}{\mu^{\pi^*}(\partial\mathcal{R}_\nu)} \in \Omega{\left( \frac{\Delta{|\mathcal{R}}|}{\Delta{|\mathcal{\partial R}}|}\right)} \;\; \forall \mathcal{R}_\nu \subset \mathcal{S}_\nu\;.\label{eq:volsurfratio}
\end{equation}
To further analyze this situation, let's consider the case where the scalable MDP is proportionally scaled among $n' \leq n$ of its $n$ continuous parameters and $n'' \leq n'$ of these parameters result in scaling of a particular region $\mathcal{R}_\nu$. To provide a scaling dependence to $\frac{\Delta{|\mathcal{R}}|}{\Delta{|\mathcal{\partial R}}|}$, we utilize the relationship between the scaling of $|\mathcal{R}_\nu|$ and the scaling of the $n''\leq n'$ intrinsic dimensions controlling its size. As in the main text, we consider the important example of proportional scaling such that the scaling of these $n''$ dimensions inherits the scaling behaviour of an $n''$-dimensional hyper-sphere with radius $\nu$. As such the state space scaling follows $|\mathcal{R}_\nu| \in \Theta(\nu^{n''})$. 
Moreover, the volume-to-surface area ratio follows $\Delta |\mathcal{R}|/\Delta |\partial \mathcal{R}| \in \Theta(\nu)$ in the case of the proportional scaling of Definition \ref{def:scalableMDP}.\footnote{This can be seen as using an n-dimensional generalization of Galileo's famous square-cube law.} This then implies that $\Delta |\mathcal{R}|/\Delta |\partial \mathcal{R}| \in \Theta(|\mathcal{S}_\nu|^{1/n'})$, since $|\mathcal{S}_\nu|\in\Theta(\nu^{n'})$ where the initial size $|\mathcal{S}_0|$ can be omitted from asymptotic notation. Finally, we can substitute this expression into equation \ref{eq:volsurfratio}: 
\begin{equation}
\frac{\mu^{\pi^*}(\mathcal{R}_\nu)}{\mu^{\pi^*}(\partial\mathcal{R}_\nu)} \in \Omega{\left( \frac{\Delta{|\mathcal{R}}|}{\Delta{|\mathcal{\partial R}}|}\right)} \in \Omega\bigg(|\mathcal{S}_\nu|^{1/n'} \bigg) \in \Omega\bigg(|\mathcal{S}_\nu|^{1/n} \bigg) \;\; \forall \mathcal{R}_\nu \subset \mathcal{S}_\nu
\end{equation}

We can then plug this result into equation \ref{eq:tmixratio}, yielding a lower bound for $t_\textrm{mix}^{\pi^*}(\epsilon)$: 

\begin{equation}
t_\textrm{mix}^{\pi^*}(\epsilon) \in \Omega\bigg(\frac{\mu^{\pi^*}(\mathcal{R}_\nu)}{\mu^{\pi^*}( \partial \mathcal{R}_\nu)}\bigg) \in \Omega{\left( \frac{\Delta{|\mathcal{R}}|}{\Delta{|\mathcal{\partial R}}|}\right)} \in \Omega\bigg(|\mathcal{S}_\nu|^{1/n}\bigg) \;\; \forall \mathcal{R}_\nu \subset \mathcal{S}_\nu : \mu^{\pi^*}(\mathcal{R}_\nu) < 1 - \epsilon
\label{eq:tmixratio}
\end{equation}

As we noted in our proof of Proposition \ref{prop:timespent}, any arbitrarily large scalable region of density greater than the bound i.e. $\mu^{\pi^*}(\mathcal{R}_\nu) \geq 1 - \epsilon$ that scales polynomially must also contain a scalable sub-region $\mathcal{R'}_\nu \subset \mathcal{R}_\nu$ of arbitrarily small density $\mu^{\pi^*}(\mathcal{R'}_\nu) < 1 - \epsilon$ that scales at least at the same rate. As such, there is no need to provide an upper bound for the steady-state region probability density in Theorem \ref{theorem:scalablemixing}. If a scalable MDP is defined, it must be that such a region exists. As we know that $n \geq n' \geq n'' \geq 1$, there must exist a scalable region $\mathcal{R}_\nu$ that has a residence time scaling with $\Omega(|\mathcal{S}_\nu|^{1/n})$ that meets our definition of a polynomial with exponent $k > 0$ for any scalable MDP with a finite description in terms of $n$ continuous variables.
The existence of a polynomial mixing time is then proven using Proposition \ref{prop:polynomialmixing}.

\section{Broader Impact Statement}

Our work focuses on highlighting the existence of polynomial mixing times as MDPs are scaled up and the resulting myopic bias for common approaches to RL in this setting. Approaches that suffer from myopic bias during learning may suffer from various sources of optimization instability that could potentially be harmful to society if experienced by agents deployed to manage high impact services and applications. For example, a myopic agent may only optimize its behavior for the short term future, even if this will result in catastrophic effects for society in the long-term. As such, addressing myopic bias resulting from polynomial mixing times is a key step towards developing reliable and trustworthy RL agents. Our work takes a first step towards this goal by outlining the foundational theory underlying the issue and we hope that future work can use these insights towards developing agents that can robustly learn to tackle large-scale real-world problems.



\end{document}